\documentclass[letterpaper]{article} %
\usepackage[submission]{aaai24}  %
\usepackage{times}  %
\usepackage{helvet}  %
\usepackage{courier}  %
\usepackage[hyphens]{url}  %
\usepackage{graphicx} %
\usepackage{natbib}  %
\usepackage{caption} %
\usepackage{supertabular}
\usepackage{longtable}  
\usepackage{tabularray}  
\usepackage[utf8]{inputenc}
\UseTblrLibrary{booktabs}

\usepackage{tabularx}
\usepackage[american]{babel}
\usepackage{color}
\usepackage{subfig}
\usepackage{amsmath,amssymb,amsfonts}
\usepackage{textcomp}
\usepackage[dvipsnames]{xcolor}
\def\BibTeX{{\rm B\kern-.05em{\sc i\kern-.025em b}\kern-.08em
    T\kern-.1667em\lower.7ex\hbox{E}\kern-.125emX}}
\usepackage{url}

\usepackage{xcolor}
\usepackage{multirow}
\usepackage{multicol}
\usepackage{microtype}
\usepackage{fixltx2e}
\usepackage{comment}
\usepackage{bm}
\usepackage{booktabs} %
\usepackage{amsmath}
\usepackage{amssymb}
\usepackage{algorithm}
\usepackage{algpseudocode}
\usepackage{amsthm}
\usepackage{nameref}

\usepackage{graphicx}
\allowdisplaybreaks
\usepackage{cleveref}

\def\onedot{.}
\def\eg{\emph{e.g}\onedot} 

\def\ie{\emph{i.e}\onedot}

\DeclareMathOperator*{\argmax}{argmax}

\newtheorem{theorem}{Theorem}

\newtheorem{definition}[theorem]{Definition}

\newlength\myindent %
\algdef{SE}[SUBALG]{Indent}{EndIndent}{}{\algorithmicend\ }%
\algtext*{Indent}
\algtext*{EndIndent}

\def\ours{\textsc{GradObstinate}}

\usepackage{newfloat}
\usepackage{listings}
\DeclareCaptionStyle{ruled}{labelfont=normalfont,labelsep=colon,strut=off} %
\floatstyle{ruled}
\newfloat{listing}{tb}{lst}{}
\floatname{listing}{Listing}
\title{Gradient-Based Word Substitution for Obstinate Adversarial Examples Generation in Language Models}

\begin{document}

\maketitle

\begin{abstract}
In this paper, we study the problem of generating obstinate (over-stability) adversarial examples by word substitution in NLP, where input text is meaningfully changed but the model’s prediction does not, even though it should. 
Previous word substitution approaches have predominantly focused on manually designed antonym-based strategies for generating obstinate adversarial examples, which hinders its application as these strategies can only find a subset of obstinate adversarial examples and require human efforts. 
To address this issue, in this paper, we introduce a novel word substitution method named \ours, a gradient-based approach that automatically generates obstinate adversarial examples without any constraints on the search space or the need for manual design principles. 
To empirically evaluate the efficacy of \ours, we conduct comprehensive experiments on five representative models (Electra, ALBERT, Roberta, DistillBERT, and CLIP) finetuned on four NLP benchmarks (SST-2, MRPC, SNLI, and SQuAD) and a language-grounding benchmark (MSCOCO). 
Extensive experiments show that our proposed \ours\ generates more powerful obstinate adversarial examples, exhibiting a higher attack success rate compared to antonym-based methods. 
Furthermore, to show the transferability of obstinate word substitutions found by \ours, we replace the words in four representative NLP benchmarks with their obstinate substitutions. 
Notably, obstinate substitutions exhibit a high success rate when transferred to other models in black-box settings, including even GPT-3 and ChatGPT. 
Examples of obstinate adversarial examples found by \ours\ are available at \url{https://huggingface.co/spaces/anonauthors/SecretLanguage}.
\end{abstract}

\section{Introduction}\label{sec:intro}

\begin{figure}[t!]
    \centering
    \includegraphics[width=0.8\columnwidth]{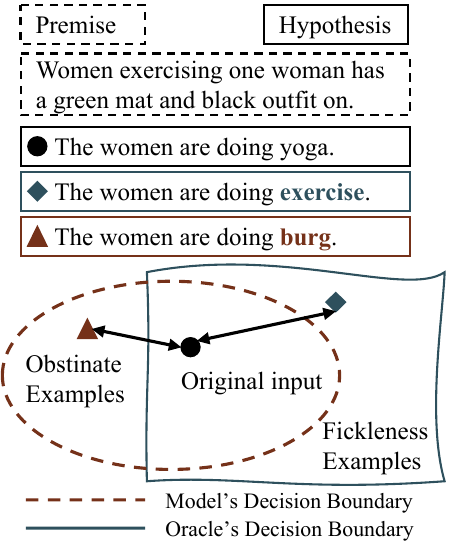}
    \vspace{-0.5em}
    \caption{
    The illustration of obstinate (over-stability) and fickle (over-sensitivity) adversarial examples. 
    We explore the generation of fickle and obstinate adversarial examples for the hypothesis ``The women are doing yoga.'' while keeping the premise the same. 
    The corresponding fickle adversarial example is ``The women are doing exercise.'', which changes the prediction of models. 
    On the other side, the obstinate adversarial example, ``The women are doing burg.'', alters the meaning of the sentence and the relationship with the premise while making the models output the same. 
    In this paper, we focus on efficiently and effectively generating obstinate adversarial examples.
    }
    \label{fig:intro}
    \vspace{-1em}
\end{figure}

In the past few years, deep learning~\cite{heDeepResidualLearning2016,devlin-etal-2019-bert} has achieved huge success, especially in the natural language processing (NLP) area. 
However, powerful deep learning methods have been shown to be easily fooled by adversarial examples, which are data crafted intentionally to confuse a model~\cite{szegedy_intriguing_2014}. 
Literature has been focused on generating and defending (fickleness) adversarial examples (see \Cref{fig:intro})~\cite{jia-liang-2017-adversarial,alzantot-etal-2018-generating,mccoy-etal-2019-right,jin_is_2020,li-etal-2020-bert-attack,DBLP:journals/corr/abs-2010-09997,alshemali-kalita-2020-generalization,zhou-etal-2021-defense} that are similar to the normal inputs with minimal perturbations and able to change the model’s output as shown in \Cref{fig:intro}. 
Such (fickleness) adversarial examples emerge when models are too sensitive around the normal data such that small changes will flip the outputs of the models. 
For example, changing ``film'' to ``movie'' might lead to a change in the output of language models (LMs)~\cite{ribeiro-etal-2018-semantically}. 
This problem~\cite{alshemali-kalita-2020-generalization,li-etal-2020-bert-attack,zhou-etal-2021-defense} in NLP has been well studied in the last few decades. 
As a remedy, synonyms and semantically similar sets can be enumerated and one can directly set supervision to force the models to learn the relationships.

On the other side, the complementary problem of generating and defending against obstinate adversarial examples (see \Cref{fig:intro}) is equally troublesome. 
Obstinate adversarial examples~\cite{niu-bansal-2018-adversarial,DBLP:conf/iclr/JacobsenBZB19,tramer_fundamental_2020,chen-etal-2022-balanced} have different semantics (true labels) compared to the original input texts while the models retain their predictions. 
Particularly, previous works can be roughly split into three categories, \ie, removing a portion of input texts~\cite{feng-etal-2018-pathologies,DBLP:conf/iclr/WelblHSGDSK20}, reordering of the texts~\cite{welbl-etal-2020-undersensitivity}, and substituting input words with other words~\cite{niu-bansal-2018-adversarial,welbl-etal-2020-undersensitivity,chen-etal-2022-balanced}. 
In this paper, we focus on the last line of literature, generating obstinate adversarial examples by word substitutions. 

To explore the vulnerability of LMs to obstinate adversarial examples generated by words substitutions, \citet{niu-bansal-2018-adversarial} first showed that adding negation or changing words with their antonyms to the original sentences will not change the outputs of several dialogue models~\cite{li-etal-2016-deep,DBLP:conf/aaai/SerbanSLCPCB17}. 
Subsequently, \citet{tramer_fundamental_2020,chen-etal-2022-balanced} showed that standard (fickle) adversarial training methods focused on reducing vulnerability to fickle adversarial examples may make models more vulnerable to obstinate adversarial examples. 
To address this issue, \citet{chen-etal-2022-balanced} introduced a novel balanced adversarial training method that incorporates contrastive learning to enhance robustness against various adversarial instances. 
However, existing methods~\cite{niu-bansal-2018-adversarial,welbl-etal-2020-undersensitivity,chen-etal-2022-balanced} in NLP only consider manual or antonym-based strategies for generating adversarial examples adhering to manually established principles, which limits the searching space and further hinders the defense to such attacks.

In this paper, to address the aforementioned challenges, we propose the first gradient-based method for generating obstinate adversarial examples in NLP, namely \ours\ (\Cref{alg:SecretFinding}). 
\ours\ automatically generates obstinate adversarial examples from the whole dictionary instead of only searching in negation and antonym sets of inputs without any human effort as opposite to \citet{niu-bansal-2018-adversarial,chen-etal-2022-balanced}. 
To avoid confusion, we establish the term "obstinate substitutions" throughout this study to denote word replacements between original sentences and their corresponding obstinate adversarial examples. 
For example, given the original hypothesis ``The women are doing \underline{yoga}'' and its corresponding obstinate adversarial example ``The women are doing \textit{burg}'', the term ``\textit{burg}'' represents an obstinate substitution for ``\underline{yoga}'', as illustrated in \Cref{fig:intro}.

To empirically evaluate our methods, we select five representative benchmarks, \ie, SST-2, MRPC, SNLI, SQuAD, and MSCOCO. 
Results show that our proposed \ours\ outperforms previous methods with the SOTA performance, proving the power of gradient-based methods. 
Then, to further understand obstinate adversarial robustness, we test the transferability of obstinate substitutions on five white-box models and two black-box models, \ie, DistillBERT, ALBERT, Roberta, Electra, GPT-2, GPT-3, and ChatGPT. 
Notably, obstinate substitutions can be easily transferred yielding consistently high success rates, which further proves the efficiency of \ours.

\medskip
\noindent
\textbf{Our contributions are summarized as follows.}
\begin{itemize}
\item 
We propose the first gradient-based method, namely \ours\ (see Algorithm~\ref{alg:SecretFinding}), for word substitution-based obstinate adversarial generation in NLP. 
Different from previous methods~\cite{niu-bansal-2018-adversarial,welbl-etal-2020-undersensitivity,chen-etal-2022-balanced}, which rely on predefined antonym-based principles for crafting obstinate adversarial examples, our method seamlessly and \textit{autonomously} generates obstinate adversarial instances \textit{without} necessitating human expertise or involvement.
\item
To validate the efficacy of our proposed \ours, we conduct comprehensive experiments on five representative NLP tasks with five language models. 
The results show that \ours\ outperforms previous methods by a large margin on five diverse tasks, which proves the superiority of our method.
\item 
To test the transferability of obstinate substitutions, we attack seven models including GPT-3 and ChatGPT with obstinate substitutions found by \ours. 
Results show that obstinate substitutions can be easily transferred to other models with high attack success rates, showing the vulnerability of LMs.
\item 
Examples of obstinate adversarial examples and substitutions found by \ours\ are available on \url{https://huggingface.co/spaces/anonauthors/SecretLanguage}. 
We will release our code upon publication.
\end{itemize}

\section{Related works}

In this section, we briefly review the literature on adversarial robustness and obstinacy in NLP. 

\medskip
\noindent
\textbf{Fickle adversarial robustness in NLP. }
(Fickle) Adversarial robustness~\cite{ren-etal-2019-generating,DBLP:journals/tist/ZhangSAL20,wang-etal-2022-measure} have been studied extensively in NLP, which aims to improve the stability of LMs towards semantically similar inputs. 
That always includes the ability to identify and use synonyms~\cite{li-etal-2020-bert-attack,morris-etal-2020-reevaluating,garg-ramakrishnan-2020-bae} and related words. 
To improve the adversarial robustness of LMs, numerous methods~\cite{gardner-etal-2020-evaluating,DBLP:conf/iclr/KaushikHL20,zhou-etal-2021-defense,DBLP:conf/aaai/SchlegelNB21} have been proposed to improve this robustness ability of NLP models. 
Some work generates fickle adversarial examples with the gradients of the victim models in character-level~\cite{ebrahimi-etal-2018-adversarial,ebrahimi-etal-2018-hotflip} or word-level~\cite{cheng_seq2sick_2020}. 
Specifically, HotFlip~\cite{ebrahimi-etal-2018-hotflip} swaps characters, based on the gradients of the one-hot input vectors while Seq2Sick~\cite{cheng_seq2sick_2020} employs the projected gradient method combined with group lasso and gradient regularization to generate word-level substitutions. 

Similar to Seq2Sick, \ours\ also uses gradients to change words, while Seq2Sick targets fickle robustness, and \ours\ focuses on obstinate robustness. 
However, the most notable difference between our proposed \ours\ and Seq2Sick is that Seq2Sick perturbs the embeddings of words, while we directly perturb the input sentences. 
Furthermore, as Seq2Sick works on embeddings, it still needs group lasso and gradient regularization to ensure that the perturbed embeddings are valid within the vocabulary. In contrast, our method does not inherently require such constraints due to the design of perturbing the sentences directly. 
Last, \ours\ employs LMs and several principles and LMs to ensure that obstinate adversarial examples avoid sharing similar semantics with the inputs, while Seq2Sick lacks a comparable design to ensure that fickle adversarial examples maintain the same semantics as the inputs. 
Hence, owing to these design nuances and distinct objectives, our method excels in efficiently and effectively generating obstinate adversarial examples compared to Seq2Sick.

\medskip
\noindent
\textbf{Obstinate adversarial robustness in NLP. }
As the complementary problem of fickle adversarial robustness, obstinate adversarial robustness has been ignored for a long time until \citet{niu-bansal-2018-adversarial} highlighted its significance by revealing the susceptibility of dialogue models to obstinate adversarial examples. 
At the same time, similar ideas have been explored by \citet{DBLP:conf/iclr/JacobsenBZB19}. 
They showed that computer vision models are stable to semantic-changing perturbations and thus make vast regions in input space vulnerable to (obstinacy) adversarial attacks. 
To address this issue, \citet{niu-bansal-2018-adversarial,welbl-etal-2020-undersensitivity,chen-etal-2022-balanced} proposed antonym-based strategies to generate fickle adversarial examples and then conduct adversarial training to improve the robustness of LMs. 
In parallel, multi-modal models have grappled with fickle adversarial examples as well.	In 2022, \citet{DBLP:journals/corr/abs-2206-00169} showed that \texttt{Apoploe vesrreaitais} means birds and \texttt{Contarra ccetnxniams luryca tanniounons} (sometimes) means bugs or pests to DALLE-2~\cite{DBLP:journals/corr/abs-2204-06125} with hand-crafted examples. 

Different from methods~\cite{niu-bansal-2018-adversarial,welbl-etal-2020-undersensitivity,chen-etal-2022-balanced,DBLP:journals/corr/abs-2206-00169} in NLP and multimodal area, which generates obstinate adversarial examples by antonym-based strategies or manually set principles, \ours\ leverages gradient methods to automatically generate obstinate adversarial examples without any human efforts and with an infinite search space. 
Benefiting from the gradient-based design, \ours\ shows superior performance in attacking 5 different language-based models compared with previous methods.

\section{Our approach}

In this section, we introduce a principled approach, namely \ours, to find obstinate adversarial examples given language models (LMs) and input sentences. 
Prior to delving into the details, we establish the necessary notations. 
Initially, we posit the existence of a human oracle $\mathcal{O}(\cdot)$, which provides the semantic interpretation of a given input. 
A sentence is represented as $s = [word_1, \ldots, word_i, \ldots, word_n]$, where $word_i$ is the $i$-th word and the length of sentence is $n$. 
The substitution of the $i$-th word in the sentence $s$ is denoted as $s^{'}_{i} = [word_1, \ldots, word_i^{'}, \ldots, word_n]$, where $word_i^{'}$ is the new word. 
The index of the target word is represented by $l \in [n]$.

\subsection{Problem overview}

Now, following \citet{niu-bansal-2018-adversarial,tramer_fundamental_2020,chen-etal-2022-balanced}, we present a formal definition of obstinate adversarial examples.
\begin{definition}
    Given an LM $f(\cdot)$, obstinate adversarial examples $s^{\prime}$ should be semantically different from and have the same outputs with the original input sentences $s$, such that,
    \begin{enumerate}
        \item \textbf{Different Semantics:} $\mathcal{O}(s^{\prime}) = \mathcal{O}(s)$.
        \item \textbf{Same Outputs:} $f(s^{\prime}) = f(s)$.
    \end{enumerate}
    Additionally, we introduce the concept of obstinate substitution, referring to the replacement of words in the original sentence, \eg, the word  ``\textit{burg}'' serving as an obstinate substitution for the word ``\underline{yoga}''.
\end{definition}

\subsection{\ours\ for finding obstinate adversarial examples}

Previous method~\cite{niu-bansal-2018-adversarial,tramer_fundamental_2020,chen-etal-2022-balanced,DBLP:journals/corr/abs-2206-00169} found obstinate adversarial examples, \eg, ``\textit{Apoploe vesrreaitais}'' as``\underline{birds}'' and ``\textit{Contarra ccetnxniams luryca tanniounons}'' as ``\underline{bugs}'' or ``\underline{pests}'', with manually-crafted sentences, which requires prior knowledge of LMs and human efforts. 
To efficiently and automatically generate obstinate adversarial examples, we propose \ours, a gradient-based algorithm detailed in \Cref{alg:SecretFinding}. It is able to return obstinate adversarial examples without any human efforts and achieves faster inference benefiting from the gradient design.

\begin{algorithm}[t!]
\caption{\ours}\label{alg:SecretFinding}
\begin{algorithmic}[1]
\Require Input sentence $s = (word_1, \ldots, word_n)$, model $f(\cdot)$, tokenizer $t(\cdot)$, target word's index $l$, step size $\eta$, step $E$. 
\Ensure Obstinate adversarial example $s^{\prime}$.
\item[]
\item[] \textbf{\ours} $(s, f, t, l, \eta, e)$:
\Indent
\State Get the one-hot representation $\mathbf{r}_{onehot} \leftarrow [\mathbf{r}_{onehot, 1}, \ldots, \mathbf{r}_{onehot, n}] \in \{0,1\}^{n \times n_{words}}$ of each word in the input sentence $s$ with the tokenizer $t(\cdot)$;
\State Initialize noise as $z \sim \mathcal{N} (0,1)$;
\For{$e\in[E]$}
\State Clip the noise as $z \leftarrow \max(\min(z, \mathbf{-1}), \mathbf{1})$;
\State Get the perturbed one-hot representation $\mathbf{r}_{onehot}^{\prime} \leftarrow [\mathbf{r}_{onehot, 1}, \ldots, \mathbf{r}_{onehot, l} + z, \ldots, \mathbf{r}_{onehot, n}]$
\State $loss \leftarrow CrossEntropy(f(\mathbf{r}_{onehot}), f(\mathbf{r}_{onehot}^{\prime}))$;
\State Update noise as $z \leftarrow z - \eta \operatorname{sign}(\frac{\partial loss}{\partial z})$;
\If {$f(\mathbf{r}_{onehot} + \mathbf{z}) == y$}
\State Break;
\EndIf
\EndFor
\State Get the index of perturbed word as $wordindex \leftarrow \argmax ([\mathbf{r}_{onehot, 1}, \ldots, \mathbf{r}_{onehot, l} + z, \ldots, \mathbf{r}_{onehot, n}])$. 
\State Get the sentences $s^{\prime}$ with the decoding function of the tokenizer;
\State \Return Obstinate adversarial example $filtering (s, s^{\prime})$.
\EndIndent
\item[]
\item[] \textbf{filtering} $(s, s^{\prime})$:
\Indent
\State Construct a synonym set $\mathcal{S}$ for $s$;
\If {$s^{\prime} \notin \mathcal{S}$}
    \If {GPT-2 thinks $s, s^{\prime}$ are not semantically similar}
        \State \Return $s^{\prime}$.
\EndIf\EndIf
\State \Return $\varnothing$.
\EndIndent
\end{algorithmic}
\end{algorithm}

Specifically, given a sentence $\mathbf{s}$, we first map words to one-hot vectors by the tokenizer of LM $f$ as $\mathbf{r}_{onehot} = [\mathbf{r}_{onehot, 1}, \ldots, \mathbf{r}_{onehot, n}] \in \{0,1\}^{n \times n_{words}}$, where $n_{words}$ is the size of word dictionary of the tokenizer. 
Next, we initialize a noise $z$ sampled from the Gaussian distribution with mean $0$ and variance $1$. 
Then, we add the noise $z$ to the one-hot representation $\mathbf{r}_{onehot}$ as $\mathbf{r}_{onehot}^{\prime} = [\mathbf{r}_{onehot, 1}, \ldots, \mathbf{r}_{onehot, l} + z, \ldots, \mathbf{r}_{onehot, n}]$. 
To make sure that the perturbed one-hot representation is valid in the vocabulary space, we clip the noise into $[-1, 1]$ as $z = \max(\min(z, \mathbf{-1}), \mathbf{1})$. 
However, we cannot ensure that it satisfies our requirements of keeping the output the same as the output of the original sentence $f(\mathbf{r}_{onehot})$. 
To address this, we employ gradient descent to update the noise vector $\mathbf{z}$ to find obstinate adversarial examples.
We measure the distance between the outputs of the modified and the original sentences using the cross-entropy loss $loss = CrossEntropy(f(\mathbf{r}_{onehot}), f(\mathbf{r}_{onehot}^{\prime}))$ and then update the noise with the sign of gradient $\frac{\partial loss}{\partial \mathbf{noise}}$. 
After $E$ predefined epochs or when the output of the modified sentence $f(\mathbf{r}_{onehot} + \mathbf{z} )$ equals the output of the original sentence  $f(\mathbf{r}_{onehot})$, we stop the updates and obtain perturbed words by applying the decoder of tokenizer to $\argmax(r_{onehot}^{\prime})$.

Our current approach to constructing obstinate adversarial examples may not consistently meet the assumption that the ground truth label of the obstinate example differs from that of the original input. 
This inconsistency arises because the impact of obstinate substitutions on the semantic meaning of the input varies depending on the specific task. 
For example, in natural language inference, changing ``the weather is great, we should go out and have fun'' to ``the weather is bad, ...'' does not affect the entailment relationship with ``we should have some outdoor activities'' since the main argument is in the second part of the sentence. 

\begin{table*}[t!]
\centering

\begin{tabular}{llll}
\toprule
Models & Tasks                     & Datasets   & Attack Targets  \\
\midrule
\midrule
Electra     &Sentiment analysis &  GLUE (SST-2) & First Sentence \\
ALBERT      &Paraphrase         &  GLUE (MRPC)  & First Sentence\\
DistillBERT & NLI                & SNLI         & Hypothesis\\
Roberta     & QA                 & SQuAD       & Question\\
\midrule
CLIP     & Image-text retrieval                & MSCOCO    & Text  \\
\bottomrule
\end{tabular}%
\vspace{-1em}
\caption{Our experimental settings on models, tasks, benchmarks, and attack targets.}\label{tab: model dataset selection}
\vspace{-1em}
\end{table*}

Then, to address this issue and ensure the sentences \ours\ exhibit distinct semantics from the input sentences, we leverage GPT-2~\cite{radford2019language} as a ``human oracle'' and a synonym-based strategy for sentence filtration. 
More specifically, we first construct a synonym set for each word using WordNet~\cite{DBLP:journals/cacm/Miller95} and subsequently verify whether the sentences fall within the synonym set. 
Additionally, in order to further assess their semantic similarity, we enlist GPT-2 by employing the prompt ``Are the following sentences semantically similar? $\{\mathbf{s}\}, \{\mathbf{s}^{\prime}$\}''. 
If $\mathbf{s}^{\prime}$ is not present within the synonym set and GPT-2 responds negatively, we treat $\mathbf{s}^{\prime}$ as an obstinate adversarial example of $\mathbf{s}$.

It is worth noting that, in comparison to previous methods~\cite{niu-bansal-2018-adversarial,tramer_fundamental_2020,cheng_seq2sick_2020,chen-etal-2022-balanced,DBLP:journals/corr/abs-2206-00169}, to ensure accurate generation of obstinate adversarial examples, we make the following special designs. 
\textbf{Expanding the search space.} The real-valued initialization is designed to ensure that the gradient can flow back to all words. In this way, the search space contains all subwords in the vocabulary. 
\textbf{Ensuring $s^{\prime}$ is valid in the vocabulary.} 
By utilizing the gradient's sign to regulate noise updates and restricting the noise within the range of $[-1,1]$, it allows us to map the perturbed one-hot representation back to words.

\section{Experiments}
We conduct experiments on five widely-used benchmarks and models to empirically evaluate our proposed \ours.

\subsection{Experimental settings}

\textbf{Benchmarks, tasks, and victim models.} 
We choose five representative language models~\cite{DBLP:journals/corr/abs-1910-01108,DBLP:conf/iclr/LanCGGSS20,DBLP:journals/corr/abs-1907-11692,DBLP:conf/iclr/ClarkLLM20,DBLP:conf/icml/RadfordKHRGASAM21}. 
These models are tested on four typical natural language processing tasks, \ie, natural language inference (NLI)~\cite{bowman-etal-2015-large}, paraphrase, question answering (QA) ~\cite{rajpurkar-etal-2016-squad}, and sentiment classification~\cite{maas-etal-2011-learning}, and one language-grounding task, image-text retrieval~\cite{linMicrosoftCOCOCommon2014}. 
The models with respect to different tasks and benchmarks are shown in Table~\ref{tab: model dataset selection}. 
The details of models and datasets are deferred to the Appendix. 
Due to the limitation of computational resources, for SST-2, MRPC, SNLI, and SQuAD, we use the first 10,000 samples from the training and testing (or validation) subsets while for MSCOCO, we only evaluate the first 1,000 samples. 

We fix premises, text, and the second sentence in the NLI, QA, and paraphrase.
Subsequently, we employ both \ours\ and the antonym-based strategy~\cite{chen-etal-2022-balanced,niu-bansal-2018-adversarial} to generate obstinate adversarial examples for hypotheses, questions, and the first sentences, respectively. In the cases of sentiment analysis and image-text retrieval, we target text inputs.
\begin{table*}[ht!]
\begin{center}
\resizebox{\textwidth}{!}{%
\begin{tabular}{l|cc|cccccc|cc}
\toprule
\multirow{3}{*}{Attack Methods} & \multicolumn{2}{c|}{Single-Sentence} & \multicolumn{6}{c|}{Multi-Sentence} & \multicolumn{2}{c}{Cross-Modal} \\
 & \multicolumn{2}{c|}{SST-2} & \multicolumn{2}{c}{MRPC} & \multicolumn{2}{c}{SNLI} & \multicolumn{2}{c|}{SQuAD} & \multicolumn{2}{c}{MSCOCO} \\
                         & Train       & Test        & Train       & Test       & Train       & Test       & Train       & Validation        & Train        & Validation        \\\midrule
Antonym-based Method     &    50.80      &      50.08       &        0.85     &      0.58      &         32.26    &     24.13       &      5.80       &      5.22       &     3.90         &      3.60       \\
Ours (\ours)             & 99.51       & 97.53       & 96.60       & 95.77      & 95.08       & 95.08      & 91.86       & 89.32       & 100.00       & 100.00      \\
$\Delta$ & {\textbf{48.71}$\uparrow$} & {\textbf{47.75$\uparrow$}} & {\textbf{95.75$\uparrow$}} & {\textbf{95.19$\uparrow$}} & {\textbf{62.82$\uparrow$}} & {\textbf{70.95$\uparrow$}} & {\textbf{86.06$\uparrow$}} & {\textbf{84.10$\uparrow$}} & {\textbf{96.10$\uparrow$}} & {\textbf{96.40$\uparrow$}}\\
\bottomrule
\end{tabular}
}
\end{center}
\vspace{-1em}
\caption{Attack success rate on SST-2, MRPC, SNLI, SQuAD, and MSCOCO when attacking the most important words selected by IG.}\label{tab: main results}
\end{table*}

\textbf{Choose the target words.} 
In order to gain deeper insights into the susceptibility of LMs, we leverage Integrated Gradients (IG; \citet{pmlr-v70-sundararajan17a}) to quantify the significance of words during model decision-making. We focus on perturbing the most important words as replacing the least important words is meaningless, which will not change the semantic meaning of the sentence with a high probability. 
For example, given the sentence ``The women are doing yoga'', replacing ``The'' with another word may have limited impact on the sentence's overall meaning. 
We defer the details of IG to Appendix. 
Different from other methods~\cite{prasad-etal-2021-extent} that use the absolute value of IG, we use the value of IG to proxy the importance of each word, as our goal of using IG is to determine the contribution of each word to the final output. 
By sorting the value of IG, we get the importance ranking of each word and choose the most important word as our target word.

\begin{figure}[t!]
    \centering
    \includegraphics[width=1\columnwidth]{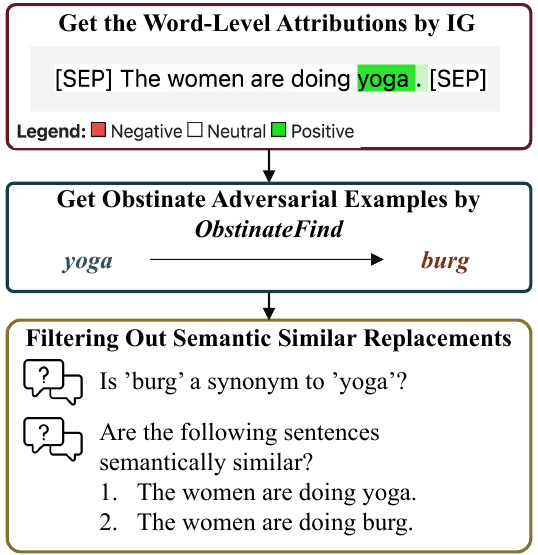}
    \vspace{-1em}
    \caption{
    The illustration of our scheme. 
    We first identify the words that are crucial to the model's decision by IG~\cite{pmlr-v70-sundararajan17a} and then utilized our proposed \ours\ to generate obstinate adversarial examples following the filtering process via synonym sets and leveraging GPT-2 as a human-like oracle.
    }
    \label{fig:scheme}
    \vspace{-1em}
\end{figure}

\textbf{Our threat model. } 
Specifically, we first identify target words that are crucial to the model's decision, and then find their substitutions with our proposed \ours\ (Algorithm~\ref{alg:SecretFinding}). 
Next, to ensure the obstinate adversarial examples are semantically dissimilar to the original sentences, following \citet{ren-etal-2019-generating}, we use GPT-2~\cite{radford2019language} and WordNet~\cite{DBLP:journals/cacm/Miller95}. 
The overall procedure is shown in Figure~\ref{fig:scheme}.

\textbf{State of the art.} 
We compare our proposed \ours\ with an antonym-based strategy~\cite{niu-bansal-2018-adversarial,chen-etal-2022-balanced}. 
We exclude \citet{welbl-etal-2020-undersensitivity} as it focuses on changing the order of the sentence instead of word substitution. 
All the experiments are run on an A100 GPU.

\textbf{Evaluation metric.} 
As our proposed \ours\ is randomly initialized and the initialization affects the success rate, we use 10 restarts for each data sample with step $E=1000$. 
We report the accuracy as the percentage of data for which at least one obstinate adversarial example can be found in 10 restarts, calculated as follows:
\begin{equation*}
    \small
    acc = \frac{\sum_{i \in [n]} \mathbf{I}( \sum_{j \in [10]}\mathbf{I}(f(x^{'}_{i, j}) == y) > 0)}{n}\,,
\end{equation*}
where $n$ is the number of data, $\mathbf{I} (cond) = 1$ when $cond$ is true, and $\mathbf{I} (cond) = 0$ otherwise. $x^{'}_{i, j}$ is the $j$-th obstinate adversarial example. 
For NLP tasks, \ie, sentiment analysis, paraphrase, NLI, and QA, $\mathbf{I}(f(x^{'}_{i, j}) == y)$ refers to the case that the outputs of original input and obstinate adversarial examples are the same, while for image-text retrieval task, that means the top-1 retrieved images of original inputs and obstinate adversarial examples are the same.

\begin{table*}[t!]
\centering

\resizebox{\textwidth}{!}{%
\begin{tabular}{m{0.4in}|m{6.5in}|m{1in}}
\toprule
\multirow{3}{*}{SST-2} & a \underline{remarkable} (\textit{improper}) 179-minute meditation on the nature of revolution.                                                      & Positive                                                    \\ \cmidrule{2-3}
                       & gangs of new york is an unapologetic \underline{mess} (\textit{dixon}), whose only saving grace is that it ends by blowing just about everything up. & Negative                                                    \\\midrule
\multirow{4}{*}{MRPC}  & \textbf{Sentence 1}: " It is safe to assume the Senate is prepared to \underline{pass} (\textit{afghanistan}) some form of cap, " King said.                 & \multirow{2}{*}{Not Equivalent}                             \\
                       & \textbf{Sentence 2}: Its safe to assume the Senate is prepared to pass some form of a cap .... The level of it is to be debated .         &                                                             \\\cmidrule{2-3}
                       & \textbf{Sentence 1}: Officials say Peeler and Jones were \underline{never} (\textit{here}) legally married but had a common-law marriage.                      & \multirow{2}{*}{Equivalent}                                 \\
                       & \textbf{Sentence 2}: Its safe to assume the Senate is prepared to pass some form of a cap .... The level of it is to be debated .         &                                                             \\\midrule
\multirow{4}{*}{SNLI}  & \textbf{Hypothesis}: An \underline{Asian} (\textit{pi}) man is surfing.                                                                                        & \multirow{2}{*}{Contradiction}                              \\
                       & \textbf{Premise}: An Asian man in a jacket, glasses, and sandals is at a high altitude aiming a gun.                                      &                                                             \\\cmidrule{2-3}
                       & \textbf{Hypothesis}: \underline{Children} (\textit{nature}) play in the snow.                                                                                  & \multirow{2}{*}{Entailment}                                 \\
                       & \textbf{Premise}: There are children playing in the snow and one of them is making a snow ball.                                           &                                                             \\\midrule
\multirow{4}{*}{SQuAD} & \textbf{Question}: What sits on \underline{top} (\textit{Chains}) of the Main Building at Notre Dame?                                                          & \multirow{2}{*}{\shortstack[l]{a golden statue \\of the Virgin Mary}} \\
                       & \textbf{Text}: Architecturally, the school has a Catholic character. Atop the Main Building .......                                       &                                                             \\\cmidrule{2-3}
                       & \textbf{Question}: Who \underline{sculpted} (\textit{volunteered}) Chopin's tombstone?                                                                         & \multirow{2}{*}{Clésinger}                          \\
                       & \textbf{Text}: Chopin's tombstone, featuring the muse of music, Euterpe, weeping over a broken lyre......                                 &                                               \\ \bottomrule      
\end{tabular}%
}
\vspace{-1em}
\caption{Examples of obstinate adversarial examples generated by \ours.
The words that our algorithm selects are marked for targets with \underline{underline}, while their obstinate substitutions are formatted as \textit{italic}.}\label{tab: quan main}
\vspace{-0.5em}
\end{table*}
\subsection{Quantitative results}

Quantitative results on five diverse models and benchmarks are shown in \Cref{tab: main results}. 
Moreover, to show the power of our proposed \ours, we replace 2-10 words in a sentence and the corresponding results are shown in \Cref{tab: acc on multisentences,tab: glue sst-2,tab: mscoco} and \Cref{fig: acc multisentence,fig: acc singlesentence} in Appendix due to the limitation of space.

\medskip
\noindent
\textbf{Quantitative results on a single-sentence task.} 
The results are shown in \Cref{tab: main results}. 
Our proposed \ours\ significantly outperforms the antonym-based strategy with a large margin on both the train and test split of SST-2. 
Specifically, \ours\ achieves remarkable attack success rates of 99.51\% and 97.54\%, whereas the antonym-based strategy only achieves a success rate slightly above fifty percent on the same dataset. 
This proves the superiority of the gradient-based method in generating obstinate adversarial examples.
Notably, as this task is much simpler than multi-sentence tasks, the success rates of both the antonym-based strategy and \ours\ are higher than those on multi-sentence tasks. 

\medskip
\noindent
\textbf{Quantitative results on three multi-sentence tasks.} 
The corresponding logits are presented in \Cref{tab: main results}. 
Similar observations can be obtained as \ours\ consistently outperforms the antonym-based method on three different benchmarks, which indicates that the proposed \ours\ is more powerful and versatile in generating effective and successful attacks across a diverse range of tasks  when compared to the antonym-based method. 
Specifically, while the antonym-based method fails on MRPC with attack success rates of only 0.85\% and 0.58\%, \ours\ demonstrates robust and consistent performance with attack success rates of  96.60\% and 96.77\%, accompanied by substantial improvements of 95.75\% and 95.19\% respectively. 
Moreover, as the task complexity escalates from paraphrase and NLI to QA, the attack success rate of \ours\ decreases from 96.60\% to 86.60\%, while the antonym-based method fails on three benchmarks with success rates less than 35\%. 

\medskip
\noindent
\textbf{Quantitative results on a cross-modal task.} 
Next, to test the generalization ability of our methods within cross-modal models, we choose CLIP and MSCOCO as our targets. 
Quantitative results on CLIP (MSCOCO) can be found in Table~\ref{tab: main results}. 
Given the nature of this task, which primarily involves the accurate retrieval of images corresponding to input sentences, its complexity stands in contrast to that of conventional NLP tasks. 
Attackers only need to find a sentence that is closer to the representation of ground-truth images than the representations of other images in the gallery. 
Remarkably, though our \ours\ achieves a 100\% attack success rate on both the train and validation sets, surpassing the antonym-based method with a huge gap as it only has 3.9\% and 3.6\% accuracy.

\subsection{Qualitative results}

To qualitatively evaluate our \ours, we present obstinate adversarial examples as shown in Table~\ref{tab: quan main}. 
More examples can be found in \Cref{tab: mrpc examples train,tab: QA examples train,tab:SNLI-train-examples} in Appendix. 

Significantly, for MPRC, SNLI, and SQuAD, obstinate adversarial examples have different semantic meanings compared to the original inputs while maintaining the same outputs. 
For instance, when we replace \underline{sculpted} with its obstinate substitution \textit{volunteered}, the meaning of the question has been changed but Roberta still outputs ``Clésinger'' with the same text.

\begin{table}[t!]
\centering
\resizebox{\columnwidth}{!}{%
\begin{tabular}{m{0.7in} |m{1.3in}|l|c}
\toprule
From                           & Obstinate Substitution  & Test on            & Success Num/ALL  \\
\midrule\midrule
\multirow{2}{*}{DistillBERT}  & \multirow{2}{*}{\underline{yoga} (\textit{burg})}     & ALBERT (GLUE (MRPC)) & 0 / 0
   \\
                               &                                  & DistillBERT (SNLI)       & 162 / 231       \\
                               &                                  & Roberta (SQuAD)        & 10 / 11       \\
                               \midrule\midrule
\multirow{3}{*}{ALBERT} & \multirow{3}{*}{\underline{scientists} (\textit{sunderland})}    & ALBERT (GLUE (MRPC)) & 8 / 8    \\
&& DistillBERT (SNLI) & 58 / 63    \\
                               &                                  & Roberta (SQuAD)      & 50 / 96 \\\midrule
\multirow{3}{*}{ALBERT} & \multirow{3}{*}{\underline{because} (\textit{resistance})}    & ALBERT (GLUE (MRPC)) & 74 / 74\\
&& DistillBERT (SNLI)       & 1,510 / 1,587     \\
                               &                                  & Roberta (SQuAD)      & 166 / 221 \\\midrule
\multirow{3}{*}{DistillBERT} & \multirow{3}{*}{\underline{people} (\textit{async})}    & ALBERT (GLUE (MRPC)) & 163 / 163    \\
&& DistillBERT (SNLI)       & 45,946 / 47,947     \\
                               &                                  & Roberta (SQuAD)      & 1,759 / 1,759 \\\midrule
\multirow{3}{*}{DistillBERT} & \multirow{3}{*}{\underline{boy} (\textit{unsett})}    & ALBERT (GLUE (MRPC)) & 25 / 25    \\
&& DistillBERT (SNLI)       & 29,782 / 32,066     \\
                               &                                  & Roberta (SQuAD)      & 69 / 69 \\\midrule
                               \multirow{3}{*}{Roberta} & \multirow{3}{*}{\underline{year} (\textit{Pref})}    & ALBERT (GLUE (MRPC)) & 460 / 460    \\
&& DistillBERT (SNLI) & 749 / 797    \\
                               &                                  & Roberta (SQuAD)      & 2,311 / 4,442 \\\midrule
\multirow{3}{*}{Roberta} & \multirow{3}{*}{\underline{50} (\textit{title})}    & ALBERT (GLUE (MRPC)) & 180 / 180    \\
&& DistillBERT (SNLI)       & 160 / 230     \\
                               &                                  & Roberta (SQuAD)      & 313 / 467 \\
\bottomrule
\end{tabular}%
}
\vspace{-1em}
\caption{Experimental results on the black-box transferability of obstinate substitutions.}
\vspace{-1em}
\label{tab: not dependent on sentences}
\end{table}

\subsection{Black-box settings via transferability}
In this part, we test the transferability of obstinate substitutions found by \ours\ and answer two questions, ``Are obstinate adversarial substitutions dependent on specific contexts?'' and ``Are obstinate adversarial substitutions dependent on specific models?''

\medskip
\noindent
\textbf{Are obstinate adversarial substitutions dependent on specific contexts?} 
We first test the example we have shown in the previous sections, ``\underline{yoga}'' (``\textit{burg}''). 
The results are shown in \Cref{tab: not dependent on sentences}.
It shows that when replacing ``yoga'' with ``burg'', the outputs of 162 samples out of 231 samples in SNLI remain unchanged. 
This suggests that ``burg'' is a universal obstinate substitution of ``yoga'' on SNLI. 
To bolster this finding, we test more obstinate substitutions, and similar observations can be obtained. 
It shows that obstinate substitutions might be independent of contexts and models consistently confuse obstinate substitutions and the original words.

\medskip
\noindent
\textbf{Are obstinate adversarial substitutions dependent on specific models?} 
To answer this question, we assess the transferability of obstinate substitutions from one model (\eg, DistillBERT) to another (\eg, Roberta). 
The corresponding results are shown in \Cref{tab: not dependent on sentences}. 
These results indicate that the obstinate substitutions are independent of models as the obstinate that we find on DistillBERT and Roberta is able to transfer to another with high attack accuracy. 
Specifically, replacing ``yoga'' with ``burg'' (found on DistillBERT/SNLI) in Roberta/SQuAD achieves a 10/11 success rate, and replacing ``scientists'' with ``sunderland'' (found on ALBERT/MRPC) in DistillBERT/SNLI yields a success rate of 58/63. 
For comprehensive experimental details, please refer to \Cref{tab: black-albert,tab: black-distillbert,tab: black-roberta} in the Appendix. 

\medskip
\noindent
\textbf{Attacking GPT-3 and ChatGPT in a black-box manner. }
As a case study, inspired by the previous results, we transfer obstinate adversarial substitutions found on GPT-2~\cite{radford2019language} and Roberta to GPT-3~\cite{DBLP:conf/nips/BrownMRSKDNSSAA20} and ChatGPT~\cite{openai_2023}. 
We randomly sample several questions from SQuAD, manually mask the most important words according to human experts, and replace the resulting questions with their obstinate adversarial substitutions found in the GPT-2 and Roberta model. 
Finally, the original question and the new question with obstinate adversarial substitutions are fed into GPT-3 with the prompt, ``Answer the following question. /n Question: \{Question\} /n Answer: '', where \{Question\} represents original and new questions. 
As for ChatGPT, we directly ask the question. 
We present the results in \Cref{tab:GPT-3}. 
We notice that, the words with higher frequency seem to be easier to be attacked in this transfer-based method, as ``amazon forest'' appears more often than ``Turner'', ``'Vernon'', and ``Gaussian integer expression'' in human language. 
This might be due to the model's bias towards popular concepts in human language. 
Besides, as ChatGPT is a variant of GPT-3.5 with reinforcement learning from human feedback, it can distinguish obstinate adversarial examples better than GPT-3 and reply with ``I can not understand what you mean''.

\begin{table}[t!]
\centering
\resizebox{\columnwidth}{!}{%
\begin{tabular}{m{4in}}
\toprule
\textbf{Question}: Which name is also used to describe the Amazon rainforest in \underline{English} (\textit{Repl})?    \\
\textbf{AO}:   {Amazon Jungle or the Amazonia} (GPT-3 and ChatGPT)                                                                   \\
\textbf{AR}:   \underline{Amazon Jungle or the Amazonia} (GPT-3 and ChatGPT)                                                                   \\
\midrule
\textbf{Question}: What method was used to clear forest for crop cultivation in the \underline{amazon forest} (\textit{am104141})?    \\
\textbf{AO}:   Slash and burn (GPT-3 and ChatGPT)                               \\
\textbf{AR}:   \underline{Slash and burn} (GPT-3 and ChatGPT)                                                                    \\
\midrule
\textbf{Question}: Who designed the \underline{Turner} (\textit{Herm}) and \underline{Vernon}(\textit{>>>>}) galleries that were built in 1858-9?            \\
\textbf{AO}:   Charles Barry (GPT-3), Sydney Smirke (ChatGPT)    \\
\textbf{AR}:   \textit{Sir Charles Barry} (GPT-3), \textit{NA} (ChatGPT)                                                                  \\
\midrule
\textbf{Question}: What do a and b represent in a \underline{Gaussian integer expression} (\textit{eless rand Blueprintued})?                        \\
\textbf{AO}: a and b represent the real and imaginary parts of the Gaussian integer expression. (GPT-3 and ChatGPT) \\
\textbf{AR}: \textit{a is the lower bound and b is the upper bound.} (GPT-3), \textit{NA} (ChatGPT)   \\
\bottomrule
\toprule
\textbf{Question}: What sits \underline{on top} (\textit{gib Logo}) of the Main Building at Notre Dame?                        \\
\textbf{AO}: Golden Dome (GPT-3 and ChatGPT) \\
\textbf{AR}: Golden Dome (GPT-3 and ChatGPT)   \\
\midrule
\textbf{Question}: \underline{What} (\textit{Brady}) is one possible serious \underline{side} (\textit{Godd}) effect of over-using antibiotics?                        \\
\textbf{AO}: kill off the good bacteria (GPT-3), development of antibiotic-resistant bacteria (ChatGPT) \\
\textbf{AR}: Antibiotic resistance (GPT-3), development of antibiotic-resistant bacteria (ChatGPT)    \\
\bottomrule
\end{tabular}%
}
\vspace{-1em}
\caption{Examples of transferring the obstinate substitutions found on GPT-2 (top-4 tables) and Roberta (bottom-2 tables) to GPT-3 and ChatGPT.
The words that our algorithm selects are marked for targets as \underline{underline}, while their obstinate substitutions are formatted as \textit{italic}.
``AO'' and ``AR'' refer to answers to the original questions and the changed questions, respectively. 
``NA'' refers to not understanding the questions.}
\label{tab:GPT-3}
\end{table}

\section{Conclusion}
In this paper, we addressed the challenge of generating obstinate adversarial examples through word substitution for LMs. 
The problem involves altering the input text while maintaining the model's prediction, despite it being expected to change. 
We proposed a novel word substitution technique, named \ours, that employed a gradient-based approach and generated obstinate adversarial examples without any human effort or predefined rules with unlimited search space. 
Extensive experiments on five diverse but representative benchmarks, \ie, SST-2, MRPC, SNLI, SQuAD, and MSCOCO, showed the superiority of \ours\ over previous methods, achieving higher attack success rates. 
Notably, obstinate substitutions generated by \ours\ exhibited significant transferability across different models, even in black-box settings on GPT-3 and ChatGPT. 
Moreover, to advance the research on the obstinacy of LMs in NLP, we have released examples of obstinate adversarial examples found by \ours\ on \url{https://huggingface.co/spaces/anonauthors/SecretLanguage} and will release our code upon publication.

In the future, as \ours\ is a gradient-based (white-box) method, it would be interesting to explore black-box methods. 
Additionally, this work only investigates the performance of \ours\ while ignoring its effectiveness in helping adversarial training. 
Moreover, as obstinate adversarial examples pose severe ethical problems~\cite{pmlr-v81-dwork18a}, such as ALBERT and DistillBERT thinking ``Asian'' equals to ``worst'' (Table~\ref{tab: black-distillbert}), it is important to figure out how to eliminate the obstinate adversarial examples and improve the performance of LMs for more transparent and interpretable LMs~\cite{saxon-etal-2021-modeling,shi-etal-2021-neural} by adversarial training or some other techniques.

\bibliography{anthology,references,custom}

\appendix

\clearpage

\begin{algorithm}[t!]
\caption{\ours, Attacking multiple words}\label{alg:SecretFinding multi words}
\begin{algorithmic}[1]
\Require Input sentence $s = (word_1, \ldots, word_n)$, model $f(\cdot)$, tokenizer $t(\cdot)$, the index of words that are intended to change $l \in [n]^{m}$, step size $\eta$, step $E$. 
\Ensure Obstinate adversarial example $s^{\prime}$.
\item[]
\item[] \textbf{\ours} $(s, f, t, l, \eta, e)$:
\Indent
\State Get the one-hot representation $\mathbf{r}_{onehot} \leftarrow [\mathbf{r}_{onehot, 1}, \ldots, \mathbf{r}_{onehot, n}] \in \{0,1\}^{n \times n_{words}}$ of each word in the input sentence $s$ with the tokenizer $t(\cdot)$;
\State Initialize noise as $\mathbf{z}_i \sim \mathcal{N} (0,1), \forall i \in \mathbf{l}$ and let $\mathbf{z}_i \leftarrow \mathbf{0}, \forall i \notin \mathbf{l}$;
\For{$e\in[E]$}
\State Clip the noise as $\mathbf{z} \leftarrow \max(\min(\mathbf{z}, \mathbf{-1}), \mathbf{1})$;
\State $loss \leftarrow CrossEntropy(f(\mathbf{r}_{onehot}), f(\mathbf{r}_{onehot} + \mathbf{z}))$;
\State Update noise as $\mathbf{z}_{i} \leftarrow \mathbf{z}_{i} - \eta \operatorname{sign}(\frac{\partial loss}{\partial \mathbf{z}_{i}}), \forall i \in \mathbf{l}$;
\If {$f(\mathbf{r}_{onehot} + \mathbf{z}) == y$}
\State Break;
\EndIf
\EndFor
\State Get the index of perturbed word as $wordindex \leftarrow \argmax (\mathbf{r}_{onehot} + \mathbf{z})$. 
\State Get the sentences $s^{\prime}$ with the decoding function of the tokenizer;
\State \Return Obstinate adversarial example $filtering (s, s^{\prime})$.
\EndIndent
\item[]
\item[] \textbf{filtering} $(s, s^{\prime})$:
\Indent
\State Construct a synonym set $\mathcal{S}$ for $s$;
\If {$s^{\prime} \notin \mathcal{S}$}
    \If {The output of GPT-2 with the input ``Are the following sentences semantically similar? $s, s^{\prime}$'' is No}
        \State \Return $s^{\prime}$.
\EndIf\EndIf
\State \Return $\varnothing$.
\EndIndent
\end{algorithmic}
\end{algorithm}

\begin{table*}[ht!]
\centering
\resizebox{\textwidth}{!}{%
\begin{tabular}{llccccccccccc}
\toprule
\multirow{2}{*}{Tasks} &\multirow{2}{*}{Datasets}     & \multirow{2}{*}{Splits} & \multicolumn{10}{c}{Number of words to be replaced}                                 \\
                             &              &          & 1     & 2     & 3     & 4     & 5     & 6     & 7     & 8     & 9     & 10    \\\midrule\midrule
\multirow{2}{*}{Paraphrase}&\multirow{2}{*}{GLUE (MRPC)} & Train                  & 99.51 & 99.29 & 98.88 & 98.31 & 97.57 & 96.59 & 95.39 & 94.47 & 92.18 & 89.72 \\
 &                            & Test                   & 97.73 & 96.58 & 96.17 & 96.34 & 95.48 & 93.62 & 93.16 & 92.46 & 90.26 & 87.88\\
                        \midrule
\multirow{2}{*}{NLI} & \multirow{2}{*}{SNLI}                  & Train                  & 95.08 & 95.38 & 93.05 & 88.00 & 81.63 & 75.63 & 70.51 & 65.42 & 62.39 & 59.47 \\
 & & Test                   & 94.51 & 95.16 & 92.96 & 89.01 & 82.05 & 76.31 & 70.68 & 65.83 & 62.57 & 59.30 \\ \midrule
\multirow{2}{*}{QA}&\multirow{2}{*}{SQuAD}   & Train                  & 91.86 & 86.68 & 80.22 & 71.63 & 69.70 & 62.20 & 48.50 & 39.20 & 31.10 & 23.40 \\
                         && Validation             & 89.32 & 82.83 & 75.00 & 65.40 & 67.20 & 55.00 & 44.30 & 34.00 & 23.30 & 18.30 \\
                             \bottomrule
\end{tabular}%
}
\caption{Accuracy (\%) of \ours\ on multi-sentence tasks when we replace various numbers of words.}
\label{tab: acc on multisentences}
\end{table*}
\begin{table*}[ht!]
\centering

\resizebox{\textwidth}{!}{%
\begin{tabular}{lccccccccccc}
\toprule
\multirow{2}{*}{Dataset}      & \multirow{2}{*}{Split} & \multicolumn{10}{c}{Number of Words Replaced}                                 \\

                              &                        & 1     & 2     & 3     & 4     & 5     & 6     & 7     & 8     & 9     & 10    \\ \midrule
\multirow{2}{*}{GLUE (SST-2)} & Train                  & 96.60 & 96.99 & 97.30 & 97.49 & 97.67 & 97.63 & 97.69 & 97.85 & 97.89 & 97.99 \\
                              & Test                   & 95.77 & 94.56 & 94.34 & 94.07 & 93.52 & 94.29 & 93.68 & 93.30 & 93.35 & 93.35\\ 
                              \bottomrule
\end{tabular}%
}
\caption{
Accuracy (\%) of \ours\ on the GLUE (SST-2) when we replace various numbers of words.}
\label{tab: glue sst-2}
\end{table*}
\begin{table*}[ht!]
\centering

\resizebox{\textwidth}{!}{%
\begin{tabular}{lccccccccccc}
\toprule
\multirow{2}{*}{Dataset}      & \multirow{2}{*}{Split} & \multicolumn{10}{c}{Number of Words Replaced}                                 \\

                              &                        & 1     & 2     & 3     & 4     & 5     & 6     & 7     & 8     & 9     & 10    \\ \midrule
\multirow{2}{*}{MSCOCO} & Train                  & 100.00 & 100.00 & 100.00 & 100.00 & 100.00 & 100.00 & 100.00 & 100.00 & 100.00 & 100.00 \\
                              & Validation                   & 100.00 & 100.00 & 100.00 & 100.00 & 100.00 & 100.00 & 100.00 & 100.00 & 100.00 & 100.00\\ 
                              \bottomrule
\end{tabular}%
}
\caption{
Accuracy (\%) of \ours\ on MSCOCO when we replace various numbers of words.}
\label{tab: mscoco}
\end{table*}

\begin{figure*}[t!]
    \centering
    \subfloat[MRPC]{\includegraphics[width=0.33\textwidth]{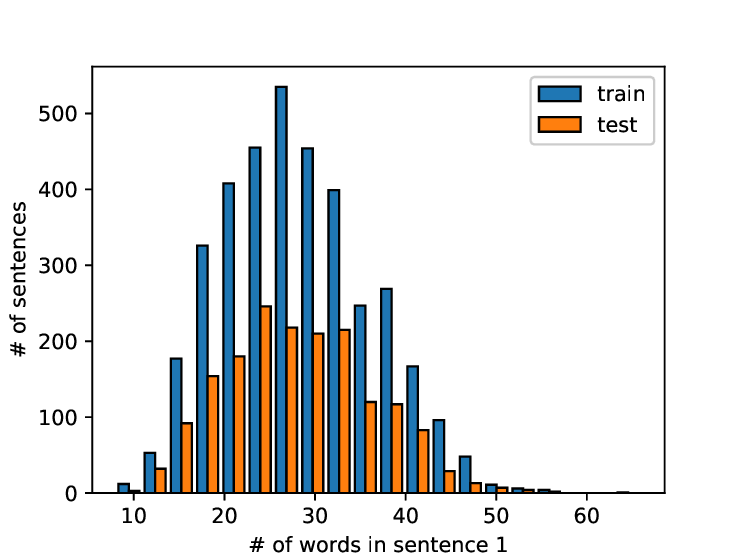}}
    \subfloat[SNLI]{\includegraphics[width=0.33\textwidth]{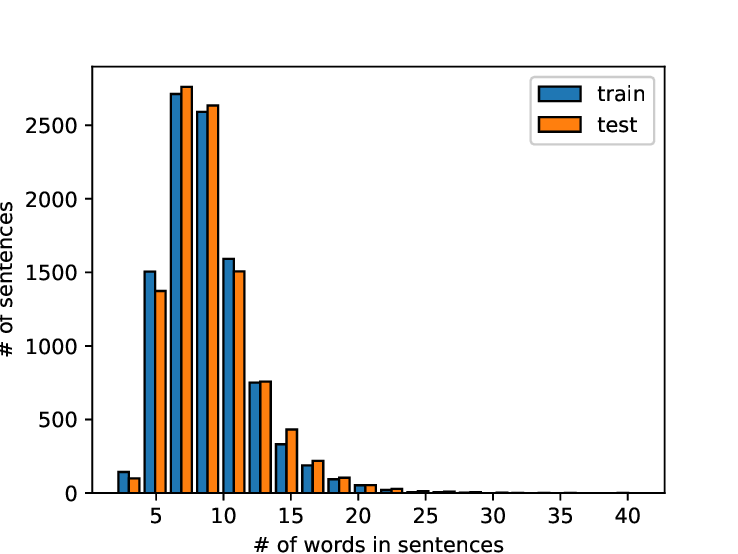}}
    \subfloat[SQuAD]{\includegraphics[width=0.33\textwidth]{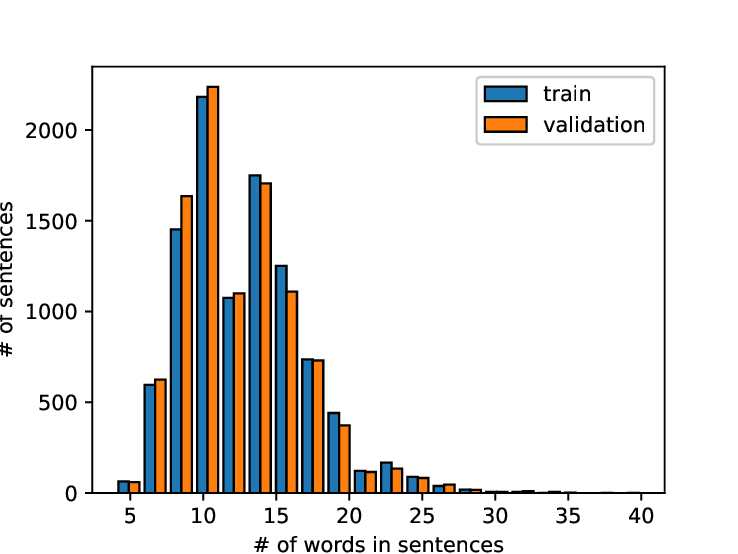}}
    \caption{The illustration of the length of first sentences, hypotheses, and questions in the train and test (validation) splits from GLUE (MRPC), SNLI, and SQuAD.
    }
    \label{fig:multi-sentences-length-hist}
\end{figure*}

\begin{figure}[t]
    \centering
    \includegraphics[width=1\columnwidth]{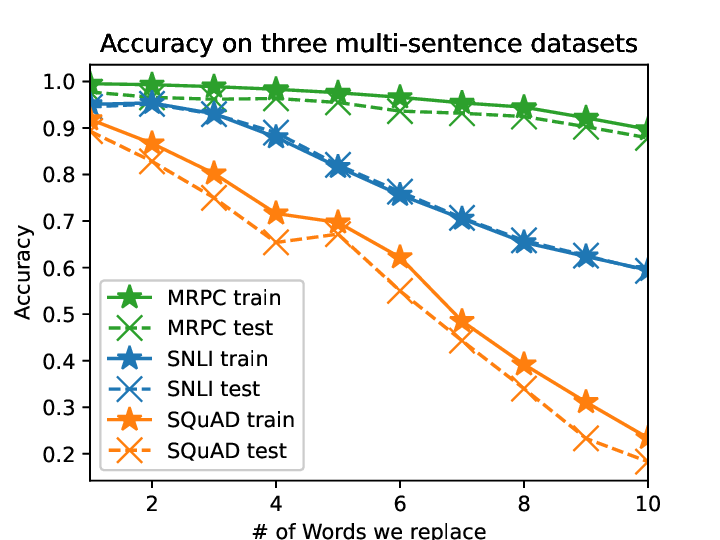}
    \caption{Accuracy of \ours\  on three multi-sentence datasets.
    }
    \label{fig: acc multisentence}
\end{figure}

\begin{figure}[t]
    \centering
    \includegraphics[width=1\columnwidth]{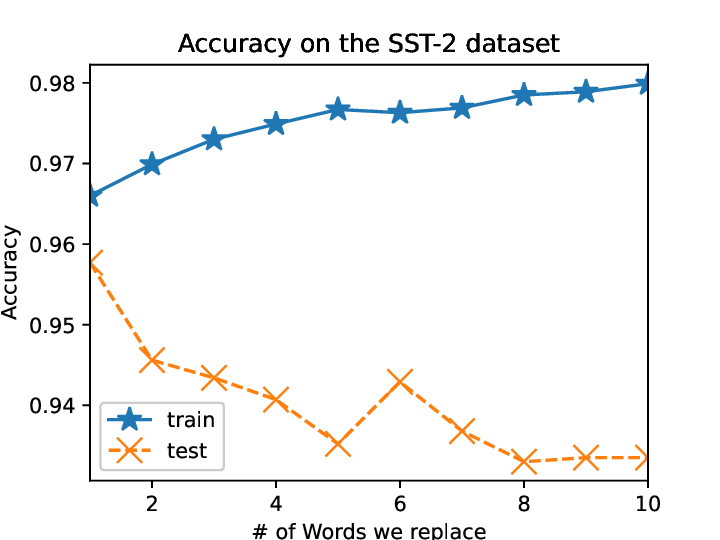}
    \caption{Accuracy of \ours\  on the single-sentence dataset (GLUE (SST-2)).
    }
    \label{fig: acc singlesentence}
\end{figure}

\section{Evaluation}

\subsection{Details of experimental settings}\label{Sec: Details of experimental settings}
\textbf{Benchmarks. }
As for the datasets, we use Stanford Natural Language Inference (SNLI,
\citet{bowman-etal-2015-large}), GLUE (MRPC)~\cite{dolan-brockett-2005-automatically}, Stanford Question Answering Dataset (SQuAD, \citet{rajpurkar-etal-2016-squad}), GLUE (SST-2)~\cite{maas-etal-2011-learning}, and MSCOCO~\cite{linMicrosoftCOCOCommon2014}. 
Stanford Natural Language Inference (SNLI, \citet{bowman-etal-2015-large}) is a collection of 570k human-written English sentence pairs manually labeled for balanced classification with the labels entailment, contradiction, and neutral. 
GLUE~\cite{maas-etal-2011-learning} is a collection of natural language understanding tasks including question answering, sentiment analysis, and textual entailment, and we use the ``MRPC'' (paraphrase) and ``SST-2'' subsets (sentiment analysis) of GLUE. 
MRPC~\cite{dolan-brockett-2005-automatically} is a corpus of sentence pairs automatically extracted from online news sources, with human annotations for whether the sentences in the pair are semantically equivalent using
SST-2~\cite{socher-etal-2013-recursive} is consisted of 67,349 training samples and 1,821 testing samples from movie reviews and human annotations of their sentiment. 
Stanford Question Answering Dataset (SQuAD, \citet{rajpurkar-etal-2016-squad}) is a reading comprehension dataset, consisting of 100,000 questions posed by crowdworkers on a set of Wikipedia articles, where the answer to every question is a segment of text, or span, from the corresponding reading passage, or the question might be unanswerable. 

For each dataset, we use the first 10,000 samples from the training and testing subsets to answer our research questions, while for SQuAD, when replacing more than 4 words, we only evaluate the first 1,000 samples due to the limitation of computational resources. 
Besides, for MSCOCO, we only evaluate the first 1,000 samples. 

Most of the single-sentence tasks and cross-modal tasks do not require a full understanding of the semantic meaning of language.  
Compared with these tasks, multi-sentence tasks are more suitable. 
Multi-sentence tasks always require models to understand the relationship between two sentences and then make predictions such as answering questions or classification results on the understanding of two sentences. 
So, when we only change one sentence while keeping another one the same\footnote{If we change one sentence while keeping another unchanged and the output keeps the same, it indicates that the relationship between two input sentences from the model side does not change and actually shows that the original sentence and the changed sentence are the same in the eye of LMs.}, it becomes a perfect testbed of our proposed method \ours. 

\subsection{Details of integrated gradient}\label{sec: IG details}
Specifically, the integrated gradient along the $i$-th dimension for an input $x$ and baseline $x^{'}$ is defined as
\begin{equation*}
    IG_i(x) = (x_i - x^{'}_i) \int_{\alpha=0}^{1}\frac{\partial f(x^{'} + \alpha (x - x^{'}))}{\partial x_i}\ d\alpha\,,
\end{equation*}
where $x_i$ and $x_i^{'}$ are the $i$-th element of $x$ and $x^{'}$.
The final integrated gradient is obtained by summing the integrated gradient along all dimensions
\begin{equation*}
    IG(x) = \sum_i IG_i(x)\,.
\end{equation*}

\begin{table*}[t!]
\centering
\resizebox{\textwidth}{!}{%
\begin{tabular}{m{5.7in}|m{1in}}%
\toprule
\textbf{Sentence1:} Media Editor Jon Friedman contributed to this story                             & \multirow{3}{*}{equivalent (train)}    \\
\textbf{Replaced Sentence1:} ashvili bo living kom caucasus shutter109eros lilly colourful          &                                                \\
\textbf{Sentence2:} Jon Friedman is media editor for CBS.MarketWatch.com in New York .              &                                                \\
\midrule
\textbf{Sentence1:} ISC and NASCAR officials declined to comment .                                  & \multirow{3}{*}{equivalent (train)}    \\
\textbf{Replaced Sentence1:} studyclo postseason burst refueling annual investing algae  auditioned &                                                \\
\textbf{Sentence2:} NASCAR officials could not be reached for comment Tuesday .                     &                                                \\
\midrule
\textbf{Sentence1:} Orange shares jumped as much as 15 percent .                                    & \multirow{3}{*}{equivalent (test)}     \\
\textbf{Replaced Sentence1:} gon consolationham enabling albeittidae drops declining fist artwork   &                                                \\
\textbf{Sentence2:} France Telecom shares dropped 3.6 percent while Orange surged 13 percent .      &                                                \\
\midrule
\textbf{Sentence1:} Total Information Awareness is now Terrorism Information Awareness .            & \multirow{3}{*}{not equivalent (test)} \\
\textbf{Replaced Sentence1:} milkggio christinaesca summonhanna finalized starecrim riley           &                                                \\
\textbf{Sentence2:} The new name will be Terrorism Information Awareness .                          &          \\         
\bottomrule
\end{tabular}%
}
\caption{Four examples from the train and test splits of GLUE (MRPC).}
\label{tab: mrpc examples train}
\end{table*}

\begin{table*}[t!]
\centering
\resizebox{\textwidth}{!}{%
\begin{tabular}{m{8in}}%
\toprule
\textbf{Text (train):} `Beyoncé announced a hiatus from her music career in January 2010, heeding her mother's advice, "to live life, to be inspired by things again". During the break she and her father parted ways as business partners. Beyoncé's musical break lasted nine months and saw her visit multiple European cities, the Great Wall of China, the Egyptian pyramids, Australia, English music festivals and various museums and ballet performances.'                             \\
\textbf{Answer:}  `her mother'                                                                                                                                                 \\
\textbf{Question:  } `Who inspired this hiatus?'                                                                                                                                  \\
\textbf{Replaced Question:   } `<s>Removeornings Ma slic consciousness'                                                                                                                     \\
\midrule
\textbf{Text (train): } `On January 7, 2012, Beyoncé gave birth to a daughter, Blue Ivy Carter, at Lenox Hill Hospital in New York under heavy security. Two days later, Jay Z released "Glory", a song dedicated to their child, on his website Lifeandtimes.com. The song detailed the couple's pregnancy struggles, including a miscarriage Beyoncé suffered before becoming pregnant with Blue Ivy. Blue Ivy's cries are included at the end of the song, and she was officially credited as "B.I.C." on it. At two days old, she became the youngest person ever to appear on a Billboard chart when "Glory" debuted on the Hot R\&B/Hip-Hop Songs chart.'               \\
\textbf{Answer:}                 `Blue Ivy Carter'             \\
\textbf{Question:}              `What was the name of Beyonce's daughter?'                                                                                                                   \\
\textbf{Replaced Question:}     `<s> arteryorio signalsahsfdceived newly tweet personalities denying'                                                                                        \\
\bottomrule
\toprule
\textbf{Text (validation):}               `Despite waiving longtime running back DeAngelo Williams and losing top wide receiver Kelvin Benjamin to a torn ACL in the preseason, the Carolina Panthers had their best regular season in franchise history, becoming the seventh team to win at least 15 regular season games since the league expanded to a 16-game schedule in 1978. Carolina started the season 14–0, not only setting franchise records for the best start and the longest single-season winning streak, but also posting the best start to a season by an NFC team in NFL history, breaking the 13–0 record previously shared with the 2009 New Orleans Saints and the 2011 Green Bay Packers. With their NFC-best 15–1 regular season record, the Panthers clinched home-field advantage throughout the NFC playoffs for the first time in franchise history. Ten players were selected to the Pro Bowl (the most in franchise history) along with eight All-Pro selections.' \\
\textbf{Answer:}              `DeAngelo Williams'   \\
\textbf{Question:}           `Which Carolina Panthers running back was waived?'    \\
\textbf{Replaced Question:}  `<s> nicotine):aband Drawingldom nut Batmandelay opposite'   \\
\midrule
\textbf{Text (validation):}                `Westwood One will carry the game throughout North America, with Kevin Harlan as play-by-play announcer, Boomer Esiason and Dan Fouts as color analysts, and James Lofton and Mark Malone as sideline reporters. Jim Gray will anchor the pre-game and halftime coverage.'    \\
\textbf{Answer:}              `Westwood One'        \\                                                                                                       \textbf{Question:}           `What radio network carried the Super Bowl?'        \\
\textbf{Replaced Question:}  `<s> TSignt inquirecairica floOU road'                                                                                                   \\   \bottomrule                                                                                                      
\end{tabular}%
}
\caption{Four examples from the train and validation splits of SQuAD.}
\label{tab: QA examples train}
\end{table*}
\begin{table*}[t!]
\centering

\resizebox{\textwidth}{!}{%
\begin{tabular}{m{1.3in} |m{6in} |m{1.3in}<{\centering}}
\toprule
Premise            & `An elderly couple dances next to a table where a boy is sitting with his head down.'       & \multirow{3}{*}{Train: \textit{contradiction}} \\
Original Hypothesis                 & `A young man has his head on the table.'                              &                                \\
Changed Hypothesis                  & `acity documenting Christina stoolicroboat FullyMarcoclintonkey'      &                                \\
\midrule
Premise                             & `A skier in electric green on the edge of a ramp made of metal bars.' & \multirow{3}{*}{Train: \underline{entailment}}    \\
Original Hypothesis                 & `The skier was on the edge of the ramp.'                              &                                \\
Changed Hypothesis & ` waterways subsistenceAGuntary on assigned fluct coincided journalistic Myrgrounds'           &                                \\
                                    \midrule
Premise                             & `A white horse is pulling a cart while a man stands and watches.'     & \multirow{3}{*}{Train: \textcolor{yellow}{neutral}}       \\
Original Hypothesis                 & `A man is watching a horse race.'                                     &                                \\
Changed Hypothesis                  & `urionSecure facets facilitates renovated philanthrop PROV clock'     &                                \\
\midrule
Premise & 'This church choir sings to the masses as they sing joyous songs from the book at a church.' & \multirow{3}{*}{Test: \textit{contradiction}} \\
Original Hypothesis      & `The church has cracks in the ceiling.'                           &                                \\
Changed Hypothesis       & ` rhySov PetraWeb caffe drains Kennedy Many'                      &                                \\
\midrule
Premise                  & `Four young girls playing in the water.'                          & \multirow{3}{*}{Test: \underline{entailment}}    \\
Original Hypothesis      & `Four girls are swimming.'                                        &                                \\
Changed Hypothesis       & ` · devices Additional Game groundwork'                           &                                \\
\midrule
Premise & `A blond-haired doctor and her African american assistant looking threw new medical manuals.' & \multirow{3}{*}{Test: \textcolor{yellow}{neutral}}       \\
Original Hypothesis      & `A doctor is studying'                                            &                                \\
Changed Hypothesis       & `ZEledge quicker rave'                                            &                                \\
\bottomrule
\end{tabular}%
}
\caption{Examples from the train and test splits of SNLI on DistillBERT.}
\label{tab:SNLI-train-examples}
\end{table*}

\begin{figure*}[h!]
    \centering
    \subfloat[Train of MRPC]{
    \includegraphics[width=0.4\columnwidth]{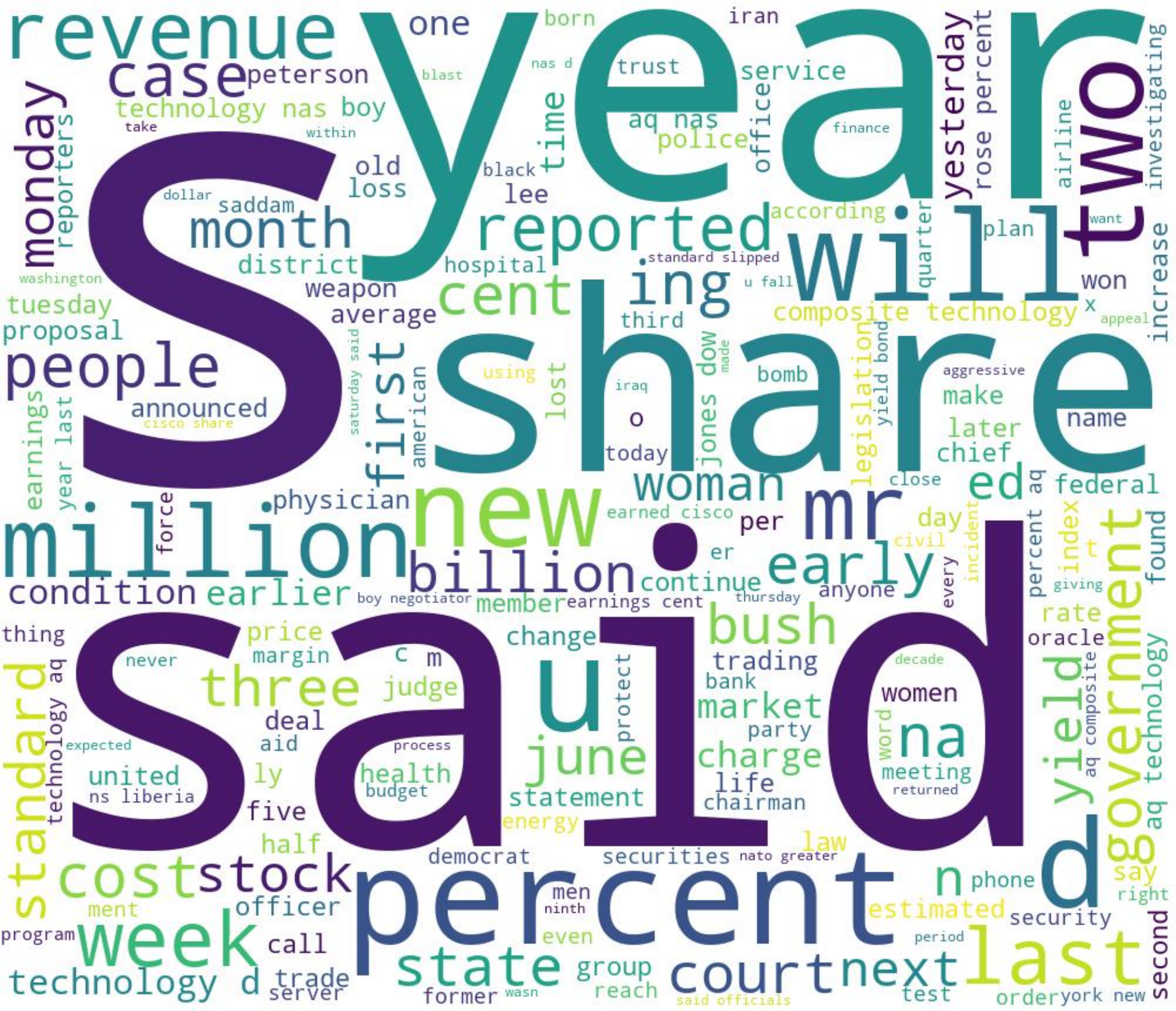}
    }
    \qquad
    \subfloat[Test of MRPC]{
    \includegraphics[width=0.4\columnwidth]{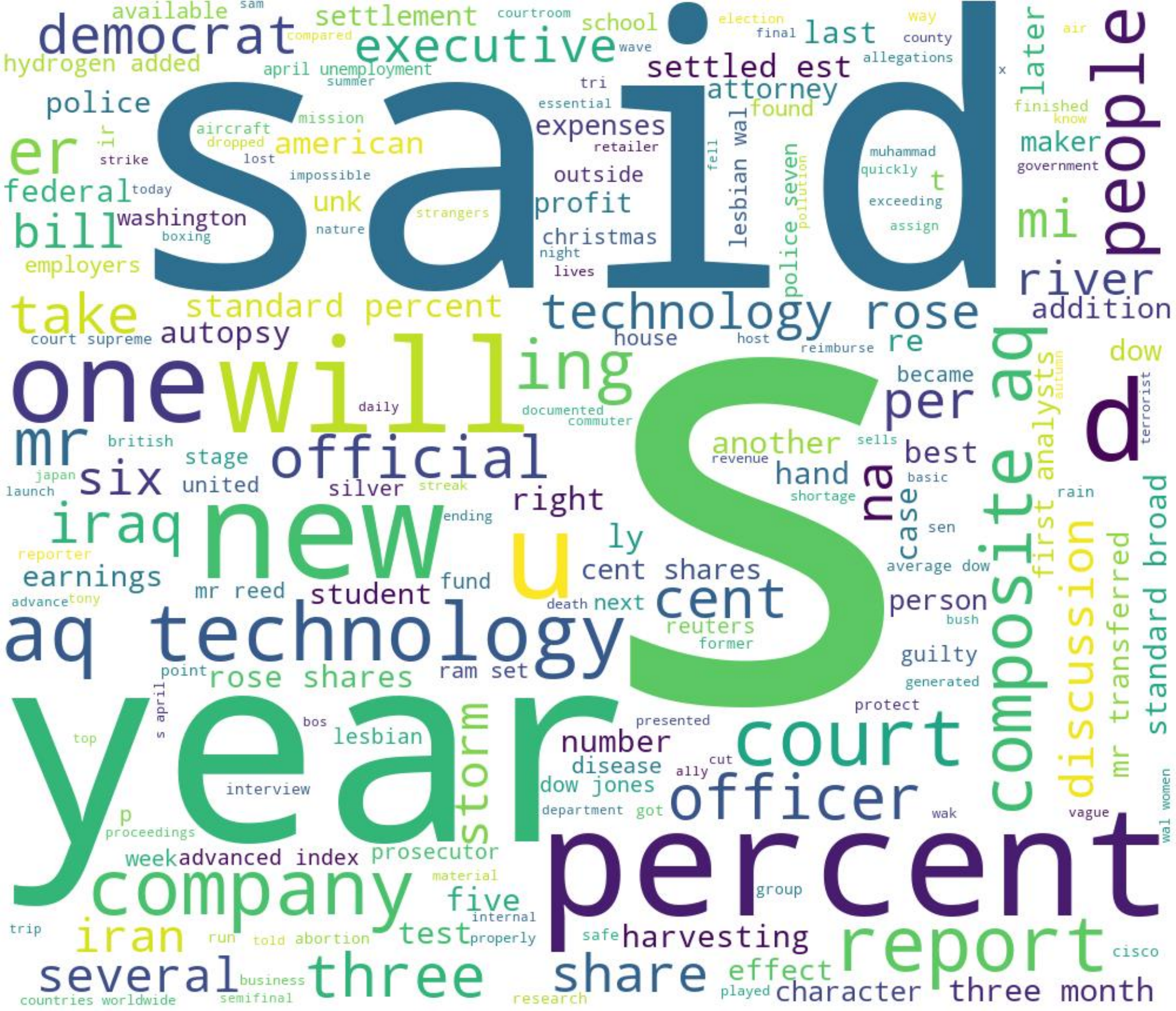}
    }
    \qquad
    \subfloat[Train of SNLI]{
    \includegraphics[width=0.4\columnwidth]{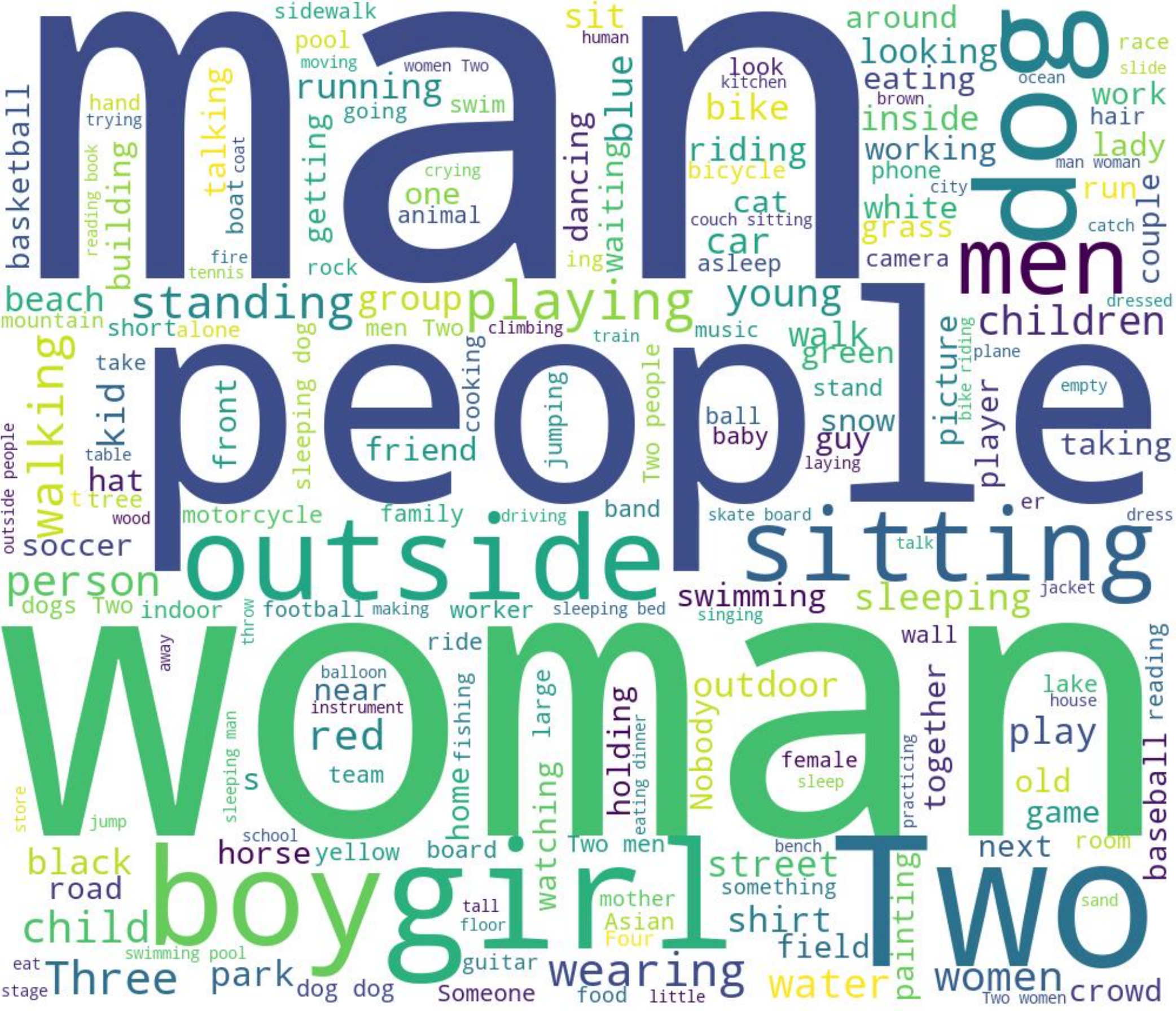}
    }
    \qquad
    \subfloat[Test of SNLI]{
    \includegraphics[width=0.4\columnwidth]{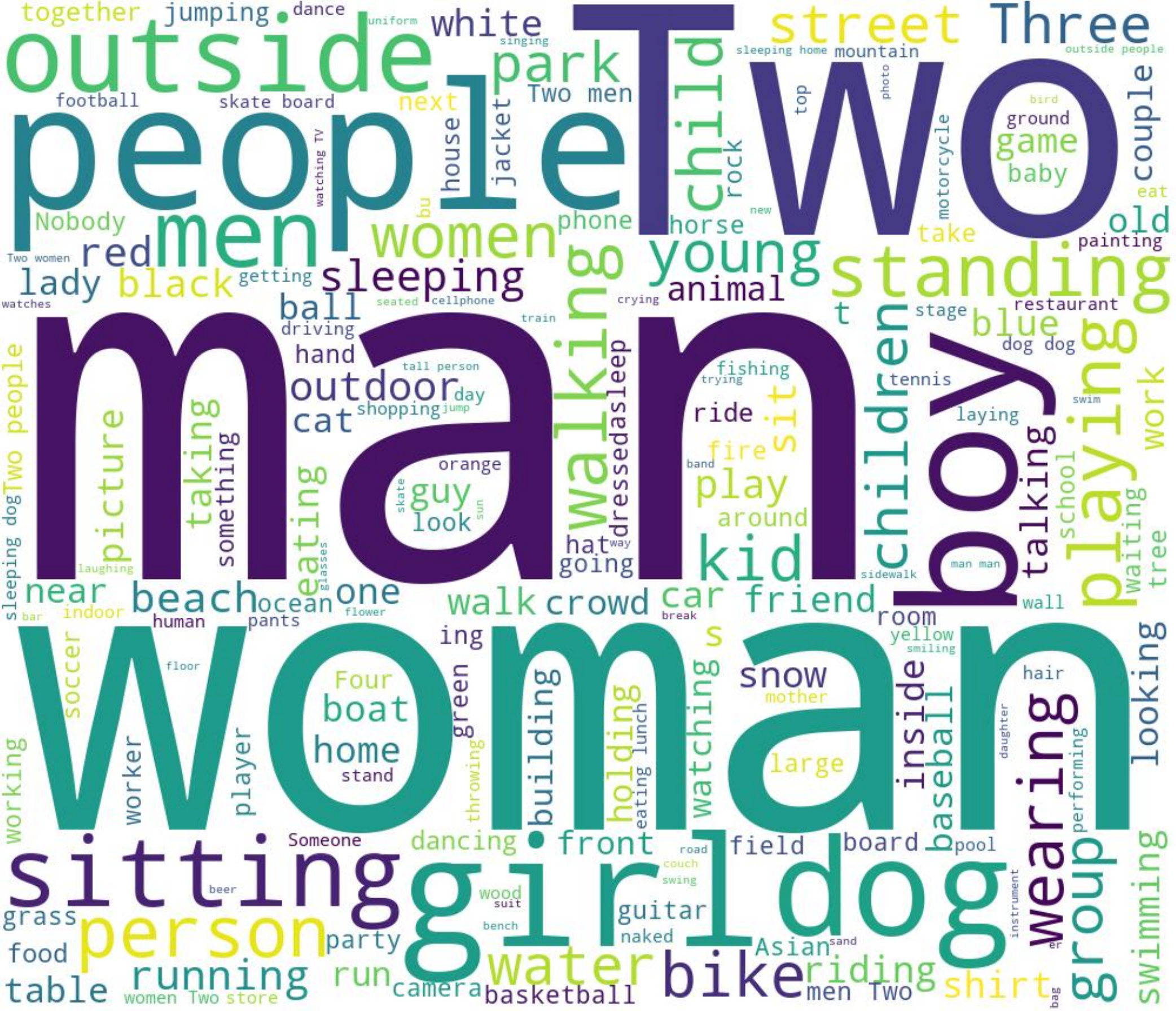}
    }
    
    \subfloat[Train of SQuAD]{
    \includegraphics[width=0.4\columnwidth]{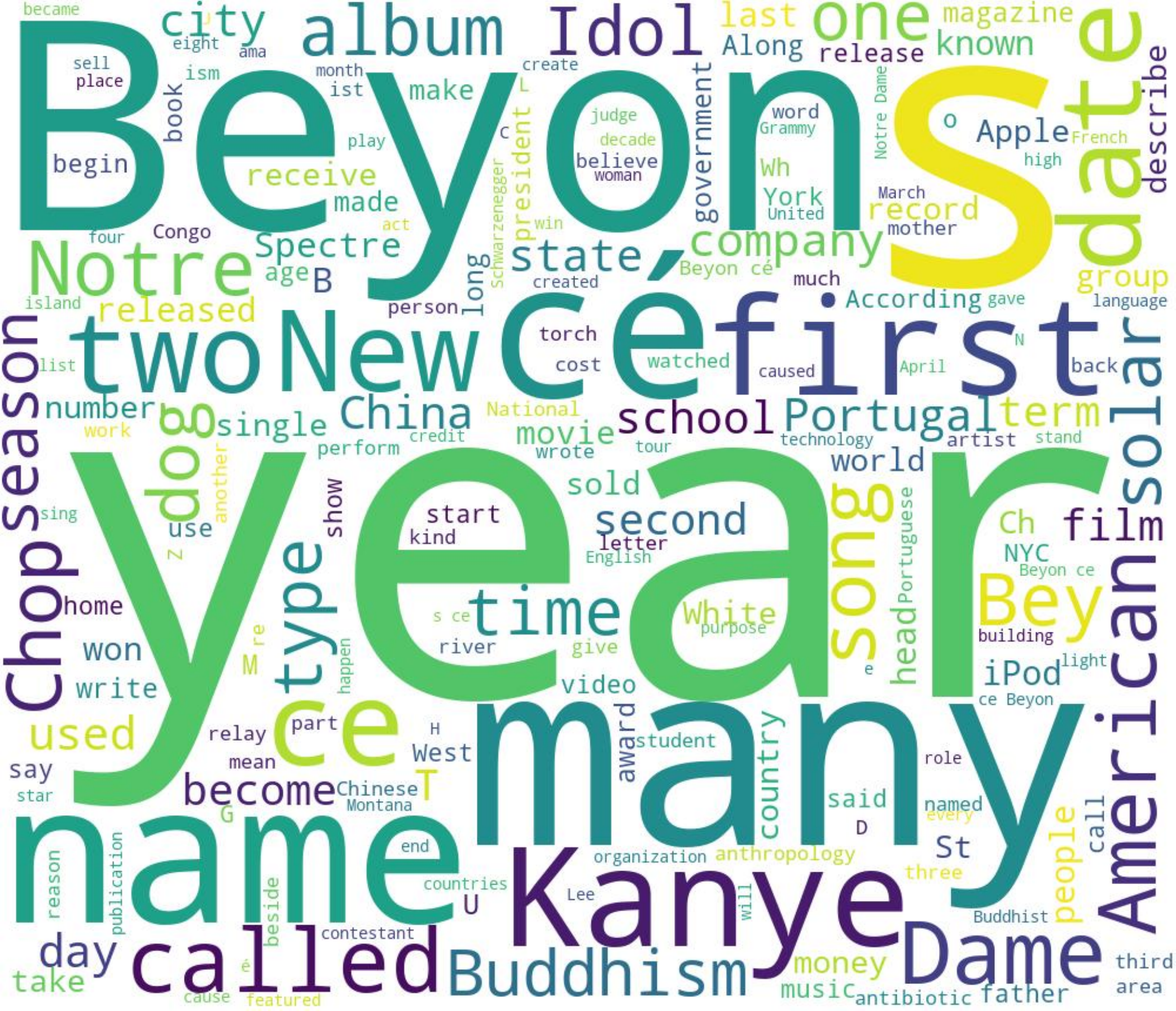}
    }
    \qquad
    \subfloat[Validation of SQuAD]{
    \includegraphics[width=0.4\columnwidth]{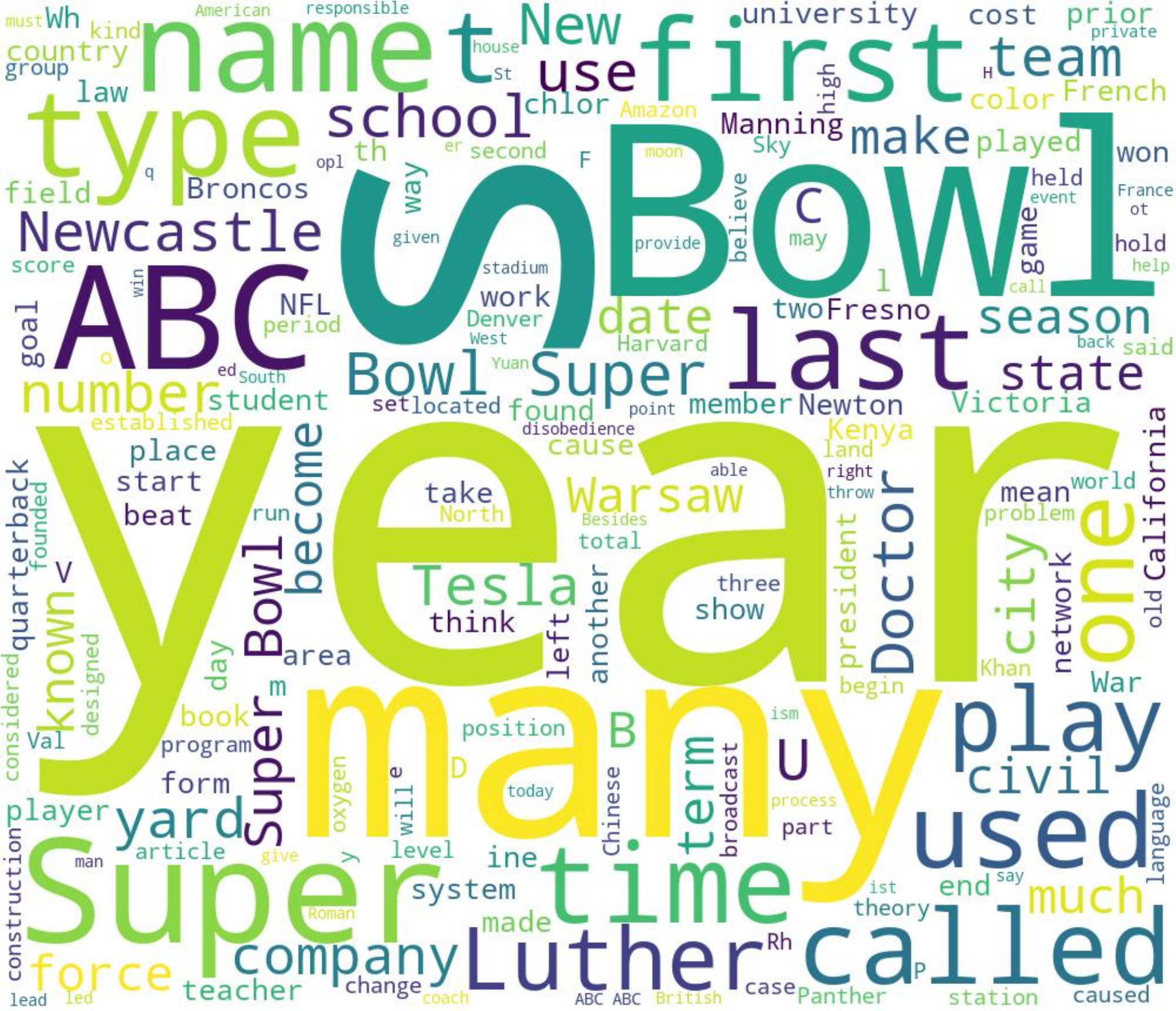}
    }
    \caption{The word clouds show the most common words selected by IG in the train and test (validation) splits of GLUE (MRPC), SNLI, and SQuAD. }
    \label{fig: word cloud on MRPC}
\end{figure*}

\begin{figure*}[t!]
    \centering
    \subfloat[Number of words of each tag on three multi-sentence benchmarks.]{
    \includegraphics[width=0.9\columnwidth]{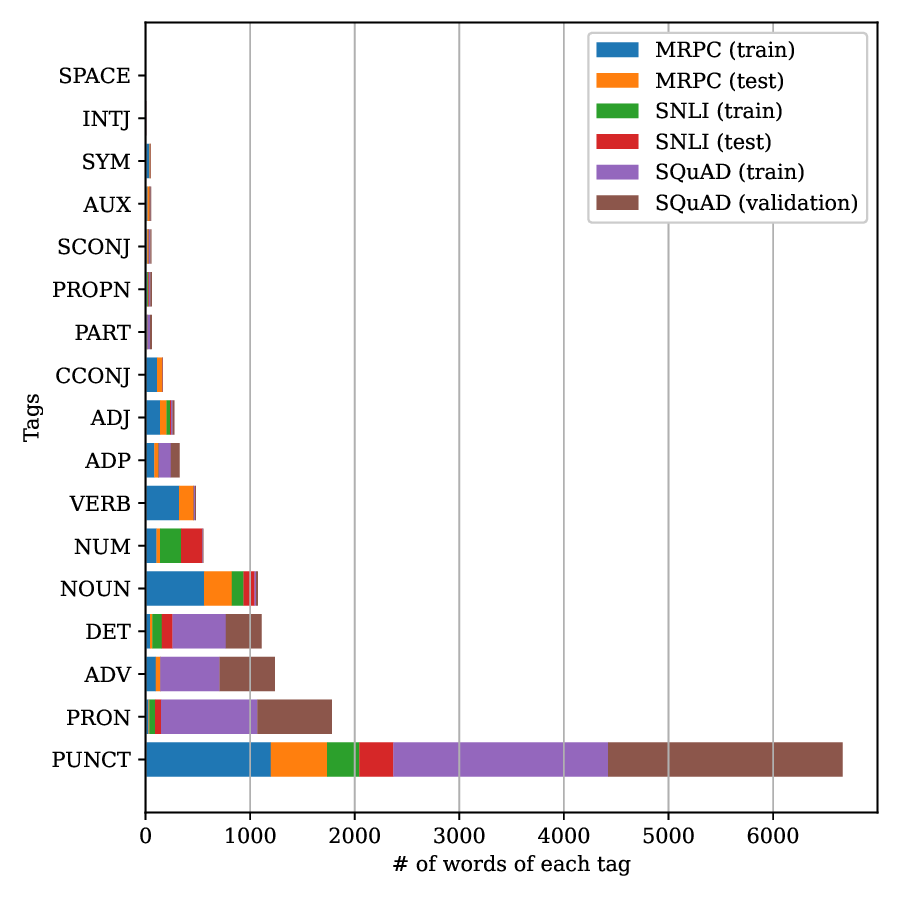}
    }
    \qquad
    \subfloat[Sankey figure showing the tag of the chosen words (left) and their obstinate substitutions (right).]{
    \includegraphics[width=0.9\columnwidth]{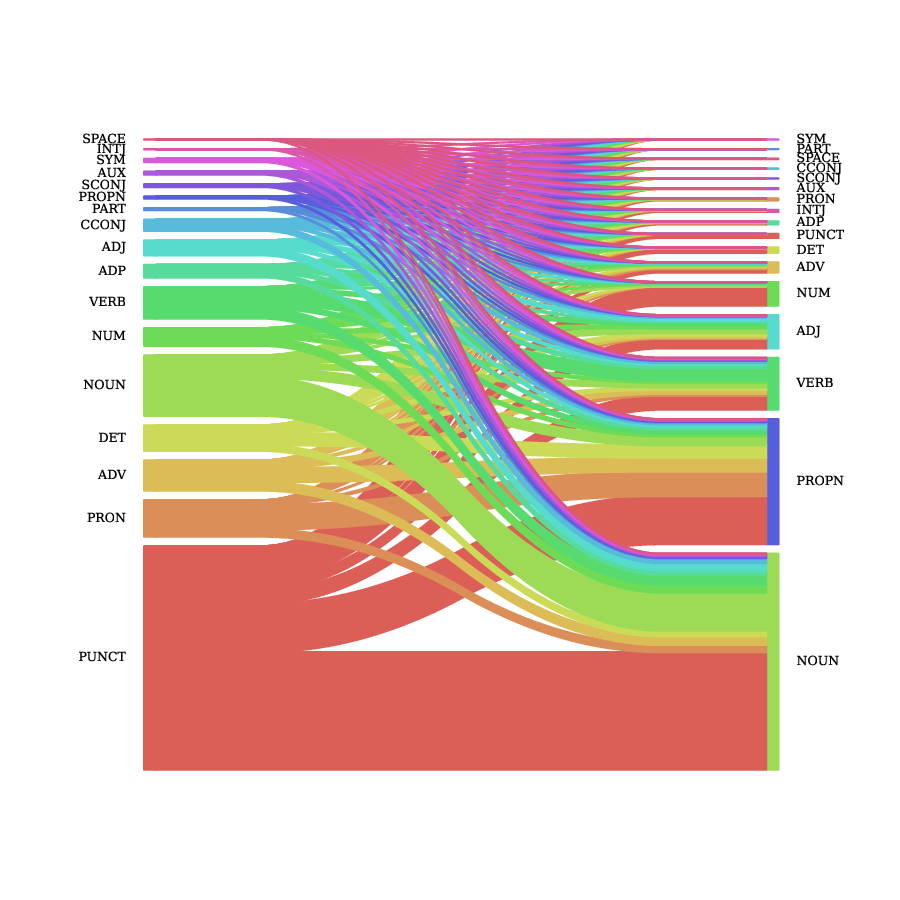}
    }
    \caption{The illustration of the tag of words.
    }
    \label{fig: multi-setence tagging num}
\end{figure*}

In the tasks of paraphrase, NLI and QA, $x$ is constructed directly by stacking all the words as $x = [CLS, A ids, SEP, SEP, B ids, SEP]$, where $CLS$ and $SEP$ are two special tokens, $A ids$ are the first sentence, question and premise, and $B ids$ are the second sentence, text, and hypothesis. 
$x^{'}$ is constructed as $x^{'} = [CLS, REFs, SEP, SEP, REFs, SEP]$, where $REFs$ is a stack of the special token $REF$ which has the same length with $Aids$ or $Bids$. 
In image-text retrieval and sentiment analysis, $x$ is constructed directly by stacking all the words as $x = [CLS, A ids, SEP]$, while $x^{'}$ is constructed as $x^{'} = [CLS, REFs, SEP]$, where $Aids$ is the text.

\subsection{Quantitative results}
To test the power of our proposed \ours, instead of only attacking one word in a sentence, we attack 2-10 words. 
The multiple-target version of \ours\ is summarized in \Cref{alg:SecretFinding multi words}. 
The corresponding results are shown in \Cref{tab: acc on multisentences,tab: glue sst-2,tab: mscoco} and \Cref{fig: acc multisentence,fig: acc singlesentence}. 

\medskip
\noindent
\textbf{Attacking more words on three multi-sentence benchmarks}. 
As expected, the accuracy for finding semantically dissimilar replacements is inversely proportional to the number of words to be replaced.
This is because the more words are changed, the more dissimilar the replacement will be compared to the original sentence.
The accuracy on the train and test splits of MRPC and SNLI achieves the highest value when 2 words are replaced, which might be due to the randomness of \ours. 
Specifically, for SNLI, the length of 7,868 and 7,758 hypotheses in the train and test splits is no more than 10. 
This means that when we change 10 words in a hypothesis, more than 77\% of the train and test split data are changed to sentences that do not share any words with the original sentence or contain any words semantically similar to words in the original sentence. 
This may be because the test (validation) and training splits are in the same domain or only the model relies on heuristics that are effective for frequent example types~\cite{mccoy-etal-2019-right}. 
And even in this case, we can still find a completely different sentence that has the same output as the original sentence, with the accuracy of 59.47\% and 59.30\% in the train and test splits, respectively.
This shows that to some extent, the model is not able to distinguish the semantic meaning of sentences. 
We show some examples of hypotheses that have been completely changed while the output remains the same in Tables~\ref{tab:SNLI-train-examples}. 
It is worth noting that in these examples, the hypotheses have been changed to completely different sentences, but the classification results remain the same and the model actually can output ``neutral'' as the \textbf{changed hypotheses are not related to the premises}.
Similar conclusions can be made on the paraphrase and QA, which demonstrates that current models, such as DistillBERT, Roberta, and ALBERT, still cannot fully understand the meaning of sentences. 

\medskip
\noindent
\textbf{Attacking more words on SST-2}. 
Quantitative results on SST-2 (Electra) can be found in \Cref{tab: glue sst-2}. 
We also draw the lines of accuracy on three multi-sentence tasks in \Cref{fig: acc singlesentence}. 
Notably, as this task is much simpler than multi-sentence tasks, the accuracy is really high and not proportional to the number of words we replace. 
Besides, the trend of accuracy on the train and test splits is different. 
As the accuracy on the train split is proportional to the number of words we replace, that on the test split is inversely proportional. 
This might be due to the fact that this task is of sentence classification and models are actually memorizing all the training data.

\medskip
\noindent
\textbf{Attacking more words on MSCOCO}. 
Quantitative results on MSCOCO (CLIP) can be found in Table~\ref{tab: mscoco}. 
As this task only requires correctly retrieving the image paired with the input sentence, it is much easier than other NLP tasks for finding obstinate substitutions.
\ours\ only needs to find a sentence that is closer to the representation of a ground-truth image than the representations of other images in the gallery. 
And as the gallery is also limited, \ours\ is able to find a replacement that is semantically dissimilar from the original sentence for every sentence. 

\subsection{Qualitative results}

\begin{table*}[t!]
\centering
\begin{tabular}{l|l}
\toprule
Words & Top 10 Frequent Obstinate Substitutions   (Train)                                  \\
\midrule
said        & tunis, machine, sichuan, 1857, quality, 81, unless, edited, sweden, his, pose   \\
for         & compatible, give, wong, relics, cius, cop, ominous, want, consent, joy          \\
that        & twi, armed, bearing, 1867, arthur, glaring, zar, halftime, progressive, autonom \\
he          & glen, female, hurting, famous, feature, radiated, jeu, 1913, hyper, petro       \\
percent     & dressed, ame, overlord, ikki, heinrich, fuji, west, paint, ossi, rash           \\
\bottomrule
\toprule
Words & Top 10 Frequent Obstinate Substitutions  (Test)                                  \\
\midrule
said        & rad, fabio, moderator, apo, gyr, education, elaine, loki, i, utter          \\
for         & monte, will, building, physically, radi, spun, bug, leiden, phar, de        \\
that        & v, back, world, marathi, violence, vers, sky, rotation, pen, sort           \\
people      & gul, accessibility, picked, bean, son, linguistic, fold, skidded, 1965, cam \\
percent     & goat, 600, mus, ramos, harmless, yu, wound, furlong, nh, raj                \\
\bottomrule
\end{tabular}%
\caption{Illustration of the top 10 most frequent obstinate substitutions of the most important words selected by IG on ALBERT from the train and test splits of GLUE (MRPC).}
\label{tab:MRPC-most-frequent-secrete-language}
\end{table*}

\begin{table*}[t!]
\centering
\begin{tabular}{l|l}
\toprule
Words & Top 10 Frequent Obstinate Substitutions    (Train)                                \\
\midrule
man         & /21955, <s>, /47119, Jewish, sym, Whereas, Ru,  consume, colon, Chief      \\
woman       & /21955, <s>, Poker, Hirosh, Vest, sodium, copies, cab, reconnect, iazep    \\
Two         & /21955, <s>, Mech, func, drama, Natural, Broadway, Lanka, casualty, aped \\
dog         & /21955, <s>, /47119, Barrel, emp, puls, basket,  pedigree, /50195,    evolved \\
people      & /21955, <s>, /47119, async, deeper, people, Copy,  tha, Looking, Charges       \\
\bottomrule
\toprule
Words & Top 10 Frequent Obstinate Substitutions  (Test)                                                             \\
\midrule
man         & /21955, <s>, /47119, /49316, Damon, Loc, peg,  009, figure, bes                              \\
woman       & /21955, <s>, uamb, checkpoints, winding, /48018, guilt, 119, po, Splash                  \\
Two         & /21955, <s>, ē, destroyer, unfolding, Celeb, unemployment, artic, Business, physical \\
dog         & /21955, <s>, /47119, hner, NI, Jr, GOP,  crispy, Assad, Previous                         \\
boy         & <s>, /21955, /47119, unsett, sensitive, ••••, rounds, stranded, lots, priced           \\
\bottomrule
\end{tabular}%
\caption{Illustration of the top 10 most frequent obstinate substitutions of the most important words selected by IG on DistillBERT from the train and test splits of SNLI.
Words ``/21955'', ``/47119'', ``/48018'', and ``/49316'' are four special words that cannot be represented in LaTeX, while 21955, 47119, 48018, and 49316 are their indices in the dictionary of the tokenizer.
``<s>'' is a special token. }
\label{tab:SNLI-most-frequent-secrete-languages}
\end{table*}

\begin{table*}[t!]
\centering

\begin{tabular}{l|l}
\toprule
Words  & Top 10 Frequent Obstinate Substitutions     (Train)                                            \\
\midrule
the    & /21955, <s>, /47119, noon, Benjamin, sshd, dataset, don, ELS, mercenary         \\
What   & /21955, /47119, <s>, Aure, frog, obsessive, excerpts, restrained, Ply, shop        \\
did    & /21955, <s>, /47119, conv, 409, adjustment, particip, afforded, organisms, band \\
year   & /21955, Pref, →, tries, /49093, picked, pedoph, lay, Restau, 647                \\
called & /21955, Enter, ud, England, hinges, iot, Geh, pay, Sem, loads      \\
\bottomrule
\toprule
Words & Top 10 Frequent Obstinate Substitutions   (Test)                                                      \\
\midrule
the   & /21955, <s>, /47119, displeasure, Ginny, encount, 950, passion, =-=-, trilogy, infiltration \\
What  & /21955, /47119, <s>, truth, rencies, Nun, VAL, Auto, Dawn, opped                           \\
did   & /21955, <s>, /47119, caravan, anny, metaph, barren, andra, officers                       \\
year  & /21955, <s>, /47119, realistic, theless, giving, Diagn, troublesome, Tomorrow, lockout    \\
50    & /21955, <s>, title, inguished, ussed, wikipedia, Robb ,Plenty, ises, commend       \\
\bottomrule
\end{tabular}%
\caption{Illustration of the top 10 frequent obstinate substitutions of the most important words selected by IG on Roberta from the train and validation split of SQuAD.
Words ``/21955'', ``/47119'', and ``/49093'' are three special words that cannot be represented in LaTeX, while 21955, 47119, and 49093 are their indices in the dictionary of the tokenizer.
``<s>'' is a special token.}
\label{tab:SQuAD-most-frequent-secrete-languages}
\end{table*}

\textbf{Examples of attacking more than one word. }
We present examples in \Cref{tab: mrpc examples train,tab: QA examples train,tab:SNLI-train-examples}. 
We notice that though most of the examples are changed totally, the output of the model remains the same. 

\medskip
\noindent
\textbf{Which words are attacked the most frequently?}
To examine whether ALBERT, DistillBERT, and Roberta consistently focus on the same words in the train and test (validation) splits, we plot the most important words according to IG in Figure~\ref{fig: word cloud on MRPC}. 
Notably, three different models perform good consistency in choosing the most important words. 
``year'', ``said'', ``share'', ``two'', and ``percent'' are frequently captured by ALBERT. 
DistillBERT shows its preference for nouns and pronouns such as ``man'', ``people'', ``dog'', and ``woman''. 
As SQuAD is mainly collected on Wikipedia, Roberta mainly focuses on the words that play the most important role in the questions, \eg, ``year'', ``name'', ``called'', ``first'', and ``school''. 
Besides, we also cluster the words we replace according to their tag in the sentence and the statistics are shown in Figure~\ref{fig: multi-setence tagging num} (a)\footnote{We use spaCy for tagging the words.}. 
It is surprising that most of the words we replace belongs to ``PUNCT'', ``PRON'', ``ADV'', and ``DET''. 
That is aligned with human behavior as humans also focus on these words too. 
But at the same time, while humans also pay attention to ``VERB'' words, models seem not to treat ``VERB'' words as the most important words in sentences. 
Next, we present the top 10 frequent obstinate substitutions of the most important words in the train and test (validation) splits in Tables~\ref{tab:MRPC-most-frequent-secrete-language}, \ref{tab:SNLI-most-frequent-secrete-languages}, and \ref{tab:SQuAD-most-frequent-secrete-languages}. 
Due to the randomness of \ours, the obstinate substitutions from the test and train splits are different while some special tokens and words consistently appear. 

\medskip
\noindent
\textbf{Which kind of words are attacked the most frequently?}
To understand the tag of their obstinate substitutions, we also draw a Sankey figure showing the tag of the most important (top 1) words by IG and their obstinate substitutions in \Cref{fig: multi-setence tagging num} (b). 
Though the tag of most of the chosen words is ``PUNCT'', notably, most of their obstinate substitutions are ``NOUN'', ``PROPN'', and ``VERB''. 
That actually corresponds with humans understanding of languages more as ``NOUN'', ``VERB'', and ``PROPN'' are always the essential components of a sentence (verb, subject, and object). 
Following ``NOUN'', ``VERB'', and ``PROPN'', a number of obstinate substitutions belong to ``ADJ'', ``NUM'', and ``ADV''. 

\subsection{Black-box settings via transferability}
In this part, we test the transferability of obstinate substitutions found by \ours\ and answer two questions, ``Are obstinate adversarial substitutions dependent on specific contexts?'' and ``Are obstinate adversarial substitutions dependent on specific models?''

To present more results on transferring obstinate substitutions to other models (black-box settings) and sentences, we randomly select 22 words, whose POS tags include 2 ``VERB'', 2 ``PRON'', 2 ``ADV'', 2 ``ADJ'', 2 ``PROPN'', 2 ``SCONJ'', 2 ``DET'', 2 ``NOUN'', 1 PUNCT'', 1 ``PART'', 1 ``AUX'', 1 ``CCONJ'', 1 ``ADP'', and 1 ``NUM'', and their corresponding obstinate substitutions from each model and dataset.
Quantitative results are shown in Tables~\ref{tab: black-albert}, \ref{tab: black-distillbert}, and \ref{tab: black-roberta}.
Generally, all the obstinate substitutions can fool ALBERT with a 100\% accuracy while that on DistillBERT is lower than that on ALBERT. 
Accuracy on Roberta is the lowest, indicating that some obstinate substitutions might not work. 
That might be because ALBERT has been finetuned on a ``simple'' task (dataset), \ie, Paraphrase (GLUE (MRPC)), compared with QA (SQuAD) and NLI (SNLI). 
And the ``simple'' task, paraphrase, does not require a deep understanding of sentences as QA and NLI do. 
So, these results actually suggest that it would be beneficial to finetune or pretrain models on large datasets and complex tasks.

At the same time, we notice that as shown in Tables \ref{tab: black-distillbert}, and \ref{tab: black-roberta}, the obstinate substituitions of ``\underline{Asian}'' and ``\underline{China}'' are ``\emph{worst}'' and ``\emph{bats}'', which raises ethical concerns about the use of language models. 

Moreover, we present the results of transferring the obstinate substitutions found on Roberta to GPT-3 and ChatGPT in Table~\ref{tab: more on ChatGPT and GPT3} to complement our black-box settings in the main paper.

\begin{table*}[ht!]
\centering

\resizebox{\textwidth}{!}{%
\begin{tabular}{m{1in}|m{1in}|m{1in}<{\centering}m{1in}<{\centering}m{1in}<{\centering}m{1in}<{\centering}m{1in}<{\centering}m{1in}<{\centering}}%
\toprule
\multirow{2}{*}{Task} & \multirow{2}{*}{Dataset} & \multicolumn{2}{c}{VERB}                    & \multicolumn{2}{c}{PRON}                      & \multicolumn{2}{c}{ADV}                  \\
                      &                          & becomes (drugged)     & challenged (zem)    & them (amphitheater) & anything (naturalist)   & simply (venomous) & almost (umu)         \\
                      \midrule
ALBERT                & GLUE (MRPC)              & 4 / 4                 & 1 / 1               & 97 / 97             & 11 / 11                 & 8 / 8             & 20 / 20              \\
DistillBERT           & SNLI                     & 8 / 11                & 7 / 7               & 1428 / 2258         & 267 / 334               & 3 / 4             & 139 / 204            \\
Roberta               & SQuAD                    & 2 / 19                & 1 / 24              & 16 / 582            & 1 / 19                  & 1 / 10            & 5 / 63               \\
\bottomrule
\toprule
\multirow{2}{*}{Task} & \multirow{2}{*}{Dataset} & \multicolumn{2}{c}{ADJ}                     & \multicolumn{2}{c}{PROPN}                     & \multicolumn{2}{c}{SCONJ}                \\
                      &                          & average (libertarian) & second (determined) & ministry (islander) & hepatitis (maclean)     & if (collectors)   & because (resistance) \\
                      \midrule
ALBERT                & GLUE (MRPC)              & 58 / 58               & 92 / 92             & 5 / 5               & 2 / 2                   & 471 / 471         & 74 / 74              \\
DistillBERT           & SNLI                     & 9 / 10                & 150 / 176           & 0 / 0               & 0 / 0                   & 6241 / 7564       & 1510 / 1587          \\
Roberta               & SQuAD                    & 117 / 327             & 283 / 593           & 1 / 15              & 1 / 1                   & 357 / 3800        & 166 / 221            \\
\bottomrule
\toprule
\multirow{2}{*}{Task} & \multirow{2}{*}{Dataset} & \multicolumn{2}{c}{DET}                     & \multicolumn{2}{c}{NOUN}                      & PUNCT             & PART                 \\
                      &                          & that (stefano)        & another (deflect)   & death (adapting)    & \textit{scientists (sunderland)} & = (appearances)   & or (machine)                    \\
                      \midrule
ALBERT                & GLUE (MRPC)              & 820 / 820             & 57 / 57             & 70 / 70             & 8 / 8                   & 30 / 30           &        3541 / 3541              \\
DistillBERT           & SNLI                     & 2879 / 3146           & 2865 / 3238         & 121 / 159           & 58 / 63                 & 3 / 3             &     84584 / 99147                 \\
Roberta               & SQuAD                    & 205 / 5007            & 500 / 588           & 260 / 405           & 50 / 96                 & 15 / 18           &     11561 / 35235                 \\
\bottomrule
\toprule
\multirow{2}{*}{Task} & \multirow{2}{*}{Dataset} & AUX                   & CCONJ               & ADP                 & NUM                     &                   &                      \\
                      &                          & have (promised)       & plus (tiger)        & alongside (caliber) & 1997 (malaysia)         &                   &                      \\
                      \midrule
ALBERT                & GLUE (MRPC)              & 372 / 372             & 10 / 10             & 3 / 3               & 7 / 7                   &                   &                      \\
DistillBERT           & SNLI                     & 2800 / 3724           & 12 / 20             & 75 / 79             & 0 / 0                   &                   &                      \\
Roberta               & SQuAD                    & 2873 / 3423           & 12 / 14             & 21 / 26             & 28 / 50                 &         \\
            \bottomrule
\end{tabular}%
}
\caption{
Experimental results (black-box) on the universality of obstinate substituitions found on ALBERT (GLUE (MRPC)). }
\label{tab: black-albert}
\end{table*}

\begin{table*}[ht!]
\centering

\resizebox{\textwidth}{!}{%
\begin{tabular}{m{1in}|m{1in}|m{1in}<{\centering}m{1in}<{\centering}m{1in}<{\centering}m{1in}<{\centering}m{1in}<{\centering}m{1in}<{\centering}}
\toprule
\multirow{2}{*}{Model} & \multirow{2}{*}{Dataset} & \multicolumn{2}{c}{VERB}            & \multicolumn{2}{c}{PRON}            & \multicolumn{2}{c}{ADV}          \\
                       &                          & training (breakthrough)              & ordering (sav)                  & Nothing (cedes) & Everyone (Speed)  & Near (IoT)           & Outside (Appearance)                  \\
                       \midrule
ALBERT                 & GLUE (MRPC)              &    9 / 9            &       1 / 1             & 0 / 0           & 1 / 1             &       6 / 6      &    1 / 1                \\
DistillBERT            & SNLI                     &    383 / 439            &        161 / 173            & 85 / 86         & 376 / 448         &       24 / 24      &      41 / 42              \\
Roberta                & SQuAD                    &   42 / 64             &        5 . 7            & 0 / 0           & 0 / 0             &      16 / 67       &           3 / 28         \\
\bottomrule
\toprule
\multirow{2}{*}{Model} & \multirow{2}{*}{Dataset} & \multicolumn{2}{c}{ADJ}             & \multicolumn{2}{c}{PROPN}           & \multicolumn{2}{c}{SCONJ}        \\
                       &                          & \textit{Asian ( worst)} & \textit{Male ( notable)}   & Billy ( Hom)    & \textit{Dog (Indonesian)}  & While (spective)           & as (Wisconsin)                  \\
                       \midrule
ALBERT                 & GLUE (MRPC)              & 5 / 5          & 4 / 4              & 2 / 2           & 2 / 2             &     18 / 18        &          2758 / 2758          \\
DistillBERT            & SNLI                     & 1279 / 1786    & 102 / 108          & 20 / 38         & 439 / 922         &     142 / 154        &      48409 / 59720              \\
Roberta                & SQuAD                    & 2 / 100        & 0 / 5              & 0 / 4           & 11 / 35           &     31 / 33        &       26014 / 34993             \\
\bottomrule
\toprule
\multirow{2}{*}{Model} & \multirow{2}{*}{Dataset} & \multicolumn{2}{c}{DET}             & \multicolumn{2}{c}{NOUN}            & PUNCT       & PART               \\
                       &                          & All (TERN)     & Some ( Nike)       & Human (Merc)    & Students ( tall)  & . (charged) & not ( settlements) \\
                       \midrule
ALBERT                 & GLUE (MRPC)              & 36 / 36        & 21 / 21            & 6 / 6           & 0 / 0             &   5792 / 5792          & 492 / 492          \\
DistillBERT            & SNLI                     & 256 / 342      & 7360 / 9592        & 525 / 538       & 152 / 212         &      435059 / 455983       & 4637 / 8410        \\
Roberta                & SQuAD                    & 32 / 166       & 7 / 142            & 8 / 94          & 0 / 9             &   1334 / 1545          & 166 / 2666         \\
\bottomrule
\toprule
\multirow{2}{*}{Model} & \multirow{2}{*}{Dataset} & AUX            & CCONJ              & ADP             & NUM               &             &                    \\
                       &                          & is (Distance)              & Neither (Imperial) & At ( Articles)  & Seven (Character) &             &                    \\
                       \midrule
ALBERT                 & GLUE (MRPC)              &   2534 / 2534             & 3 / 3              & 109 / 109       & 9 / 9             &             &                    \\
DistillBERT            & SNLI                     &    251073 / 263000            & 15 / 29            & 287 / 303       & 117 / 125         &             &                    \\
Roberta                & SQuAD                    &    14157 / 35280            & 0 / 0              & 19 / 699        & 2 / 42            &             &                \\
\bottomrule
\end{tabular}%
}
\caption{
Experimental results (black-box) on the universality of obstinate substituitions found on DistillBERT (SNLI). 
}
\label{tab: black-distillbert}
\end{table*}

\begin{table*}[ht!]
\centering

\resizebox{\textwidth}{!}{%
\begin{tabular}{m{1in}|m{1in}|m{1in}<{\centering}m{1in}<{\centering}m{1in}<{\centering}m{1in}<{\centering}m{1in}<{\centering}m{1in}<{\centering}}%
\toprule
\multirow{2}{*}{Task} & \multirow{2}{*}{Dataset} & \multicolumn{2}{c}{VERB}                & \multicolumn{2}{c}{PRON}          & \multicolumn{2}{c}{ADV}             \\
                      &                          & Using (sequent)     & wrote (aturation) & What (chieve)     & Her (nat)     & When (father) & Typically (Ezekiel) \\
                      \midrule
ALBERT                & GLUE (MRPC)              & 1 / 1               & 14 / 14           & 7 / 7             & 28 / 28       & 15 / 15       & 1 / 1               \\
DistillBERT           & SNLI                     & 3 / 3               & 51 / 54           & 4 / 6             & 123 / 137     & 10 / 13       & 0 / 0               \\
Roberta               & SQuAD                    & 10 / 12             & 11 / 376          & 39356 / 42319     & 31 / 108      & 3796 / 6115   & 0 / 4               \\
\bottomrule
\toprule
\multirow{2}{*}{Task} & \multirow{2}{*}{Dataset} & \multicolumn{2}{c}{ADJ}                 & \multicolumn{2}{c}{PROPN}         & \multicolumn{2}{c}{SCONJ}           \\
                      &                          & \textit{American (culation)} & Natural (tex)     & September (esson) & \textit{China (bats)}  & Once (ATURES) & Because (Christian) \\
                      \midrule
ALBERT                & GLUE (MRPC)              & 108 / 108           & 1 / 1             & 33 / 33           & 22 / 22       & 4 / 4         & 4 / 4               \\
DistillBERT           & SNLI                     & 449 / 518           & 1 / 2             & 5 / 5             & 51 / 143      & 2 / 4         & 2 / 2               \\
Roberta               & SQuAD                    & 49 / 1020           & 15 / 17           & 3 / 74            & 151 / 370     & 10 / 12       & 36 / 41             \\
\bottomrule
\toprule
\multirow{2}{*}{Task} & \multirow{2}{*}{Dataset} & \multicolumn{2}{c}{DET}                 & \multicolumn{2}{c}{NOUN}          & PUNCT         & PART                \\
                      &                          & All (culture)       & Each (igroup)     & All (culture)     & Each (igroup) & ? (coni)      & 's (ession)         \\
                      \midrule
ALBERT                & GLUE (MRPC)              & 36 / 36             & 3 / 3             & 36 / 36           & 3 / 3         &       12 / 12        & 1077 / 1077         \\
DistillBERT           & SNLI                     & 266 / 342           & 23 / 25           & 266 / 342         & 23 / 25       &      16 / 20         & 6635 / 6876         \\
Roberta               & SQuAD                    & 72 / 166            & 14 / 15           & 72 / 166          & 14 / 15       &       87127 / 97048        & 7765 / 9163         \\
\bottomrule
\toprule
Task                  & Dataset                  & AUX                 & CCONJ             & ADP               & NUM           &               &                     \\
                      &                          & Was (alia)          & while (Copenhagen)                 & Against (EO)      & Three (Aim)   &               &                     \\
                      \midrule
ALBERT                & GLUE (MRPC)              & 58 / 58             &       88 / 88            & 6 / 6             & 15 / 15       &               &                     \\
DistillBERT           & SNLI                     & 31 / 34             &         8557 / 10292          & 3 / 3             & 7522 / 7906   &               &                     \\
Roberta               & SQuAD                    & 76 / 321            &         123 / 161          & 11 / 18           & 24 / 40       &               &          \\          
\bottomrule
\end{tabular}%
}
\caption{
Experimental results (black-box) on the universality of obstinate substituitions found on Roberta (SQuAD). 
}
\label{tab: black-roberta}
\end{table*}
\begin{table*}[ht!]
\centering

\resizebox{\textwidth}{!}{%
\begin{tabular}{lll}
\toprule
Questions                                                            & GPT-3                                                   & ChatGPT                                          \\
\midrule
Which \underline{program} at Notre Dame \underline{offers} a Master of Education degree?     & the College of Education                                & a Master of Education                            \\
Which \textit{Mand} at Notre Dame\textit{atives} a Master of Education degree?         & NA                                                      & a Master of Education                            \\
\midrule
When did Beyoncé \underline{rise to} fame?                                       & NA                                                      & 1990s                                            \\
When did Beyoncé\textit{uk benefic} fame?                                     & Beyoncéuk benefic fame in the late 1990s as lead singer & 1990s                                            \\
\midrule
When \underline{was} the \underline{Luria}-Delbruck experiment?                              & 1943                                                    & 1943                                             \\
When\textit{age} the\textit{drawnuria}-Delbruck experiment?                            & 1940                                                    & NA                                               \\
\midrule
\underline{When} did antibiotics revolutinzed medicine?                          & 1940                                                    & 1928                                             \\
\textit{isively} did antibiotics revolutinzed medicine?                        & Yes                                                     & 1940s                                            \\
\bottomrule
\end{tabular}%
}
\caption{Examples of transferring the obstinate substituitions of Roberta to GPT-3 and ChatGPT (Dec 15 Version).
The words that our algorithm selects for substitution are marked as \underline{blue}, while their obstinate substituitions are marked as \textit{red}.
``NA'' refers to not understanding the questions.}
\label{tab: more on ChatGPT and GPT3}
\end{table*}

\end{document}